\documentclass[10pt,twocolumn,letterpaper]{article}

\usepackage[pagenumbers]{cvpr} 

\usepackage[dvipsnames]{xcolor}

\usepackage{graphicx}
\usepackage{gensymb}
\usepackage{multirow}
\usepackage{makecell}
\usepackage{tabularx}
\usepackage{booktabs}

\usepackage{amsmath}
\usepackage{amssymb}

\usepackage{url}

\usepackage[accsupp]{axessibility} 

\newcommand{\norm}[1]{\left\lVert#1\right\rVert}

\def\rvc{{\mathbf{c}}}

\def\rvh{{\mathbf{h}}}

\def\rvo{{\mathbf{o}}}

\def\rvx{{\mathbf{x}}}

\def\rvz{{\mathbf{z}}}

\def\rmI{{\mathbf{I}}}

\def\rmP{{\mathbf{P}}}

\def\gL{{\mathcal{L}}}

\def\gV{{\mathcal{V}}}

\def\sE{{\mathbb{E}}}

\newcommand{\papername}{MorpheuS}

\definecolor{cvprblue}{rgb}{0.21,0.49,0.74}
\usepackage[pagebackref,breaklinks,colorlinks,citecolor=cvprblue]{hyperref}

\def\paperID{} 
\def\confName{CVPR}
\def\confYear{2024}

\title{\papername{}: Neural Dynamic 360$\degree{}$ Surface Reconstruction \\from Monocular RGB-D Video }

\author{Hengyi Wang \quad Jingwen Wang \quad Lourdes Agapito\\
Department of Computer Science, University College London\\
{\tt\small \{hengyi.wang.21, jingwen.wang.17, l.agapito\}@ucl.ac.uk}
}

\begin{document}
\twocolumn[{%
    \renewcommand\twocolumn[1][]{#1}%
    \maketitle
    \centering
    \vspace{-0.95cm}
    \includegraphics[width=0.96\linewidth]{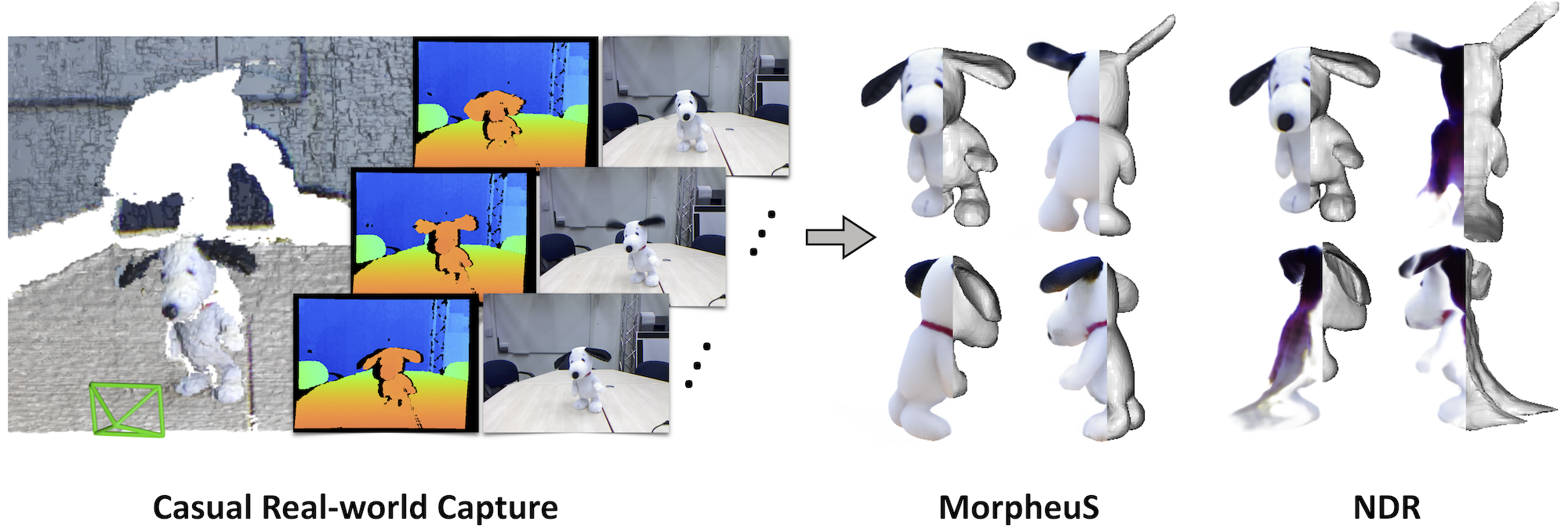}
\vspace{-3pt}
\captionof{figure}{We propose \papername{}, a dynamic scene reconstruction method that leverages neural implicit representations and diffusion priors for achieving 360$\degree{}$ reconstruction of a moving object from a monocular RGB-D video. Our approach can achieve both metrically accurate reconstruction of the observed regions and photo-realistic completion of unobserved regions of a dynamic scene.}
    \vspace{4mm}
}]

\begin{abstract}
Neural rendering has demonstrated remarkable success in dynamic scene reconstruction. 
Thanks to the expressiveness of neural representations, prior works can accurately capture the motion and achieve high-fidelity reconstruction of the target object.
Despite this, real-world video scenarios often feature large unobserved regions where neural representations struggle to achieve realistic completion.
To tackle this challenge, we introduce \papername{}, a framework for dynamic 360$\degree{}$ surface reconstruction from a casually captured RGB-D video.
Our approach models the target scene as a canonical field that encodes its geometry and appearance, in conjunction with a deformation field that warps points from the current frame to the canonical space. 
We leverage a view-dependent diffusion prior and distill knowledge from it to achieve realistic completion of unobserved regions.
Experimental results on various real-world and synthetic datasets show that our method can achieve high-fidelity 360$\degree{}$ surface reconstruction of a deformable object from a monocular RGB-D video.
Project page: \url{https://hengyiwang.github.io/projects/morpheus}.

\end{abstract}
\section{Introduction}
The challenge of reconstructing the dense 3D geometry, appearance, and motion of dynamic scenes from videos has persisted for decades, offering a wide range of applications in virtual reality and augmented reality.
Traditional dense non-rigid reconstruction methods often assume strong priors on the object's shape and motion due to the inherent ambiguities of the problem~\cite{xiao2004closed,ozden2004reconstructing}.
%
Model-based approaches~\cite{weise2011realtime, Alldieck_2018_CVPR, Kolotouros_2019_ICCV, cao_2014, Thies_2016_CVPR, kanazawa2018learning} require 3D parametric models~\cite{anguelov2005scape, SMPL:2015, palafox2021npms, blanz1999morphable, li2017learning, zuffi20173d} of specific object categories such as humans~\cite{anguelov2005scape, SMPL:2015}, faces~\cite{blanz1999morphable, li2017learning}, or animals~\cite{zuffi20173d}, resulting in promising but category-constrained reconstructions.
In this paper, we focus on model-agnostic approaches that could generalize to generic non-rigid objects. This task is often approached by jointly fusing a canonical shape and a per-frame deformation field from an RGB-D video sequence~\cite{newcombe2015dynamicfusion, innmann2016volumedeform, slavcheva2017killingfusion, Slavcheva_2018_CVPR, bozic2020deepdeform, bozic2020neural}. 
Due to the constraints of physical sensors and the reliance on traditional voxel-grid representations, they often struggle to produce high-fidelity surface details and can only reconstruct observed regions.

Recent attention has shifted towards leveraging neural representations for scene reconstruction. Neural Radiance Fields (NeRF)~\cite{mildenhall2020nerf} represent scene density and color in the weights of a neural network. In combination with volume rendering, NeRF achieves unprecedented performance on novel view synthesis.
Many follow-up works adapt this idea into surface reconstruction~\cite{yariv2021volume,wang2021neus,azinovic2022neural,wang2022go}, SLAM~\cite{sucar2021imap,zhu2022nice,wang2023co,kong2023vmap}, dynamic scene reconstruction~\cite{pumarola2021d,park2021nerfies, park2021hypernerf,Cai2022NDR} and more. 
Notably, NDR~\cite{Cai2022NDR} adopts NeRF-style representations for dynamic RGB-D surface reconstruction. Thanks to the expressiveness of neural representations, they achieve accurate and smooth surface reconstruction.
However, real-world captures often contain large unobserved regions. The inherent smoothness of neural representations, and the difficulty to incorporate domain-specific priors, cause current methods to fall short of achieving realistic completion.

In this paper, we present \papername, a framework for neural dynamic 360$\degree{}$ surface reconstruction from casually captured monocular RGB-D videos. 
We represent the target object in a hyper-dimensional canonical field and adopt a deformation field to deform the target object from observation space to hyper-dimensional canonical space, in similar spirit to HyperNeRF~\cite{park2021hypernerf}. 
In terms of reconstruction of the unobserved regions, inspired by DreamFusion~\cite{poole2022dreamfusion}, we employ a diffusion prior, i.e. Zero-1-to-3~\cite{liu2023zero1to3}, and perform Score Distillation Sampling~\cite{poole2022dreamfusion} (SDS) to distill knowledge from the diffusion prior to complete the unobserved geometry and appearance of the target object. 

Our key contribution is to demonstrate the capability to learn metrically accurate geometry and deformations of dynamic objects from casually captured RGB-D videos while achieving realistic completion in unobserved regions with diffusion priors.
We propose a temporal view-dependent SDS strategy to improve the realism of completion while learning an accurate deformation field.
A canonical space regularization strategy is used to avoid the trivial solution of surface completion.
To the best of our knowledge, MorpheuS is the first to achieve accurate, photo-realistic 360$\degree$ surface reconstruction of an arbitrary dynamic object from casually captured monocular RGB-D video. 
\section{Related Work}

\noindent
\textbf{Dense Non-rigid Reconstruction.} Traditional dense dynamic scene reconstruction methods often require strong prior knowledge about the object shape and motion, fitting 3D prior parametric models~\cite{SMPL:2015, romero2017embodied, zuffi20173d, li2017learning, weise2011realtime, Alldieck_2018_CVPR, Kolotouros_2019_ICCV, cao_2014, Thies_2016_CVPR, kanazawa2018learning} to video sequences of a specific object category, and thus cannot work with arbitrary unseen objects.
Early model-agnostic methods either require accurate dense correspondences \cite{garg2013dense} or a template shape of the object to be reconstructed first before tracking the dynamic motion~\cite{suwajanakorn2014total, zollhofer2014real, yu2015direct, bartoli2015shape}.
The seminal work DynamicFusion~\cite{newcombe2015dynamicfusion} was the first direct and template-free method to achieve real-time simultaneous tracking and reconstruction of generic non-rigid objects via joint optimization of a canonical shape and a per-frame deformation field.
Follow-up methods improve the robustness of motion tracking and reconstruction quality by incorporating photo-metric constraints~\cite{guo2017real}, classical ~\cite{innmann2016volumedeform} or learned correspondences~\cite{bozic2020deepdeform, bozic2020neural}, 
reformulating the optimization as direct dense SDF alignment~\cite{slavcheva2017killingfusion, Slavcheva_2018_CVPR}.
Although these methods have shown promising reconstruction results, they fail to complete the unobserved regions due to their inherent traditional scene representation. 

\noindent
\textbf{Neural Rendering for Dynamic Scenes.} The success of NeRF~\cite{mildenhall2020nerf} in novel view synthesis has inspired many follow-up works~\cite{pumarola2021d,park2021nerfies,li2021neural,park2021hypernerf,gao2021dynamic,tretschk2021non,wang2021dct,fang2022fast,Cai2022NDR,gao2022monocular,yang2022banmo,yang2023reconstructing,li2023dynibar,song2023total,mohamed2023dynamicsurf,mihajlovic2024resfields} in dynamic scene reconstruction. These approaches can be broadly classified into two paradigms: 
\textbf{a)} Directly modeling scene geometry and motion as a unified 4D space-time field in world (observation) space~\cite{li2021neural,gao2021dynamic,cao2023hexplane,fridovich2023kplane,li2023dynibar,shao2023tensor4d,isik2023humanrf}, where motion priors, such as scene flow or trajectory can be involved as an additional regularization~\cite{li2021neural,wang2021dct,li2023dynibar}. 
\textbf{b)} Learning a canonical shape by warping the shape from observation space to canonical space via a separate deformation field~\cite{pumarola2021d, park2021nerfies, park2021hypernerf, tretschk2021non, fang2022fast, Cai2022NDR, gao2022monocular, weng2022humannerf,yang2022banmo,yang2023reconstructing,song2023total,mohamed2023dynamicsurf, choe2023spacetime}. 
Among these works, NDR~\cite{Cai2022NDR} is particularly relevant to our work. They achieve high-fidelity dynamic surface reconstruction with neural representations. However, the continuity of neural representations does not allow realistic completion in unobserved regions.

\noindent
\textbf{Neural Rendering with Diffusion Priors.} Recent years have witnessed remarkable progress in diffusion-based foundation models~\cite{rombach2022stable,saharia2022imagen}, unlocking unprecedented capabilities for diverse applications such as image synthesis, editing, and shape generation~\cite{po2023state}.
DreamFusion~\cite{poole2022dreamfusion} is a notable example that leverages pre-trained image diffusion priors to optimize a NeRF model through SDS, achieving high-quality text-to-3D generation. 
Building upon DreamFusion's success, numerous follow-up works have emerged, particularly in text-to-3D~\cite{lin2023magic3d,chen2023fantasia3d,metzer2023latent,wang2023score,wang2023prolificdreamer,zhu2023hifa}, 
image-to-3D~\cite{melas2023realfusion,xu2023neurallift,qian2023magic123,tang2023dreamgaussian},
scene editing~\cite{haque2023instruct,shao2023control4d,pan2023avatarstudio}, etc.
Some other works~\cite{liu2023zero1to3,shi2023mvdream,liu2023syncdreamer,kong2024eschernet} try to improve the generation quality via fine-tuning pre-trained diffusion models. 
A noteworthy example is Zero-1-to-3~\cite{liu2023zero1to3}, which fine-tunes stable-diffusion~\cite{rombach2022stable} with relative pose conditioning.
%
%

\noindent
\textbf{Concurrent works.} Recently, several works have been proposed for dynamic scene generation~\cite{singer2023mav3d,jiang2023consistent4d,ren2023dreamgaussian4d,ling2023align,yin20234dgen,pan2024fast,wang2024dpdy} and editing~\cite{shao2023control4d,pan2023avatarstudio}.
MAV3D~\cite{singer2023mav3d} adopts a text-to-video diffusion model~\cite{singer2022makeavideo} for text-to-3D generation.
Control4D~\cite{shao2023control4d} utilizes a 2D diffusion prior to train a 4D GAN for 4D editing of human portraits. 
AvatarStudio~\cite{pan2023avatarstudio} fine-tunes the diffusion prior with time stamps and viewpoints, enabling editing of human head avatars with a view-and-time aware SDS loss.
Consistent4d~\cite{jiang2023consistent4d}, Efficient4D~\cite{pan2024fast}, and DpDy~\cite{wang2024dpdy} use diffusion prior for RGB-based dynamic scene reconstruction.
Different from these works, our focus is learning metrically accurate motion and geometry from monocular RGB-D video while achieving realistic completion in the unobserved region with a diffusion prior.

\begin{figure*}[h]
    \centering
    \includegraphics[width=0.98\textwidth]{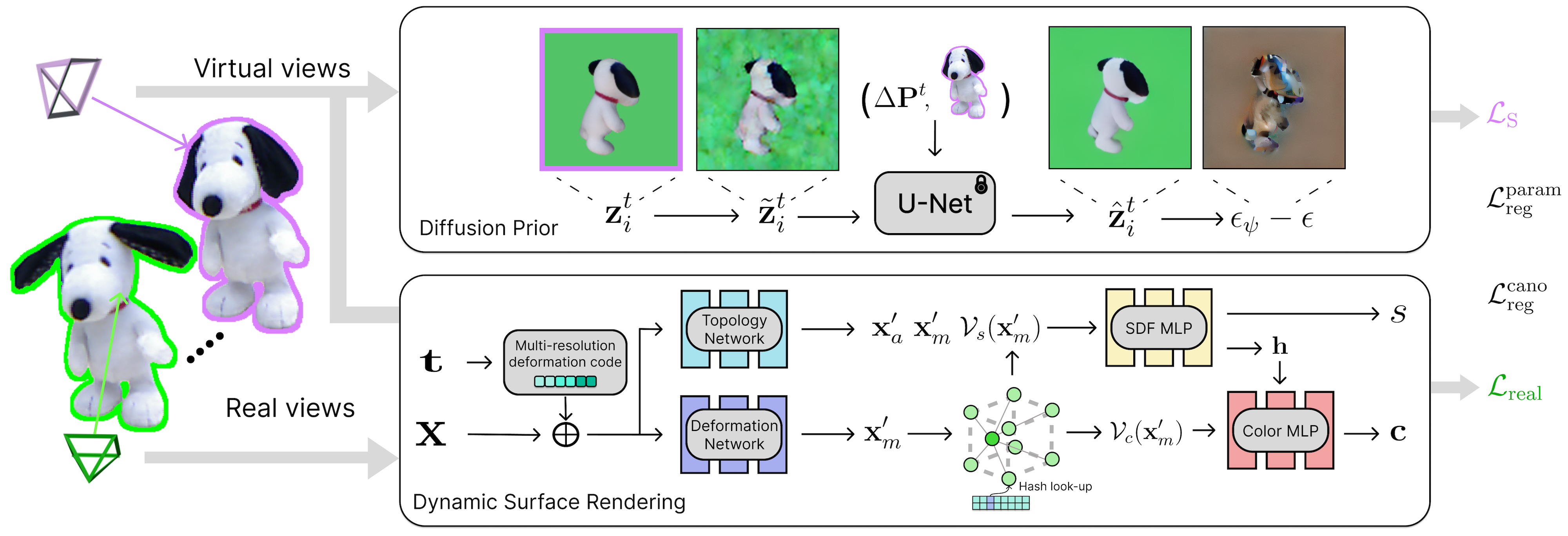}
    \caption{\textbf{Overview of \papername{}.} 1) Dynamic surface rendering: we model the target dynamic scene via a deformation field that maps a point from observation space to a hyper-dimensional canonical space and a canonical field that decodes the point into SDF and color. 2) Diffusion prior: we leverage a diffusion prior and perform SDS to complete the unobserved region. Note here the de-noising process is in latent space. All visualization is generated via decoding the latent vector for illustration purposes. 3) Optimization: We optimize the scene representation using real view supervision $\gL_{\mathrm{real}}$, SDS loss $\gL_{\mathrm{S}}$, canonical regularization $\gL_{\mathrm{reg}}^{\mathrm{cano}}$, and parameter regularization $\gL_{\mathrm{reg}}^{\mathrm{param}}$.}
    \label{fig:overview}
    \vspace{-5pt}
\end{figure*}

\section{Method}

The goal of \papername{} is to achieve 360$\degree{}$ reconstruction of an arbitrary dynamic object from a casual monocular RGB-D video $\{I_t\}_{t=1}^N, \{D_t\}_{t=1}^N$ with known camera intrinsics $K\in \mathbb{R}^{3 \times 3}$ and camera poses $P_t \in \mathbb{SE}(3)$. 
Fig.~\ref{fig:overview} shows the overview of \papername{} pipeline. 
The target scene is represented as a hyper-dimensional canonical field and a deformation field (Sec.~\ref{sec:DNSR}). 
The unobserved region reconstruction is addressed through the distillation of knowledge from a diffusion prior (Sec.~\ref{sec:DP}). 
We use both real-view observation and diffusion prior to produce supervision for our scene representation. 
Several regularization terms are used to improve the robustness of the optimization process (Sec.~\ref{sec:opt}).

\subsection{Dynamic Neural Surface Rendering}\label{sec:DNSR}

\textbf{Deformation Field.} Inspired by the recent success of dynamic scene reconstruction~\cite{Cai2022NDR}, we model the deformation of the scene via a deformation network $D(\cdot)$ and a topology network $T(\cdot)$. 
Given a point $\rvx_t \in \mathbb{R}^3$ in observation space at frame $t$, $D(\cdot)$ and $T(\cdot)$ map the point from the observation space to a point in the hyper-dimensional canonical space $\rvx^\prime \in \mathbb{R}^{3+m}$:
\begin{equation}
    \begin{aligned}
    \rvx^\prime=\{\rvx^\prime_{m}, \rvx^\prime_{a}\} = \{\rvx_t + D(
    \phi(\rvx_t), \gV_t(t)), T(\phi(\rvx_t), \gV_t(t))\},\\
    \end{aligned}
\end{equation}

\noindent
where $\rvx^\prime_m \in \mathbb{R}^3$ is the 3D coordinate of the deformed point and $\rvx^\prime_a \in \mathbb{R}^m$ is the ambient coordinate predicted by the topology network $T(\cdot)$. 
$\phi(\cdot)$ and $\gV_t(\cdot)$ are encoding functions of the location $\rvx_t$ and time $t$. 
We use a frequency encoding $\phi(\cdot)$ to model the position $\rvx_t$ for simplicity. 
For $\gV_t(\cdot)$, we use a multi-resolution 1-D feature grid instead of a per-frame deformation code~\cite{park2021hypernerf,Cai2022NDR}. 
The feature vector for each frame $t$ is obtained via linear interpolation.

\noindent
\textbf{Hyper-dimensional Canonical Field.} The 3D geometry and appearance of the target object are represented as the zero-level set of the signed distance function (SDF) field $s$ and color $\rvc$:
\begin{align}
    s, \rvh &= f_\gamma(\gV_s(\rvx^\prime_m), \rvx^{\prime}_{a}),\\
    \rvc &= f_\alpha(\gV_c(\rvx^\prime_m), \rvh),
\end{align}

\noindent
where $\gV_s$ and $\gV_c$ represent the encoding of the input position $\rvx^\prime_m$. 
Here we choose the Hash encoding~\cite{mueller2022instant} for the fast optimization speed. 
To incorporate the diffusion prior, a good initialization is important for the canonical field. 
Existing works in text-to-3D tasks often choose to initialize the radiance field with a fixed blob function~\cite{poole2022dreamfusion, lin2023magic3d}.
However, we find that the deformation field fails to converge with the fixed blob function as an initialization.
Thus, we adopt the geometric initialization following SAL~\cite{atzmon2020sal}. 
Since it is not straightforward to map the Hash features with ambient coordinates into a sphere, we append the 3D coordinate $\rvx^\prime_m$ and use zero initialization to effectively mask out the remaining coordinates in the input layer of the decoder $f_\gamma(\cdot)$.

\noindent
\textbf{Volume Rendering.} We transform the SDF into volume density $\sigma_\beta(s)$ following VolSDF~\cite{yariv2021volume}, where $\beta$ is a learn-able parameter which controls the smoothness of the surface. The volume density is then used for color $C$ and depth $D$ rendering:
\begin{equation}
\begin{gathered}
C =\sum_{i=1}^N T_i \alpha_i \mathbf{c}_i, \quad D = \sum_{i=1}^N T_i \alpha_i d_i, \quad T_i=\prod_{j=1}^{i-1}\left(1-\alpha_j\right).
\end{gathered}
\end{equation}

\noindent 
Here $T_i$ represents transmittance. The alpha value $\alpha_i$ is calculated as $1-\exp (-\sigma_i \delta_i)$, where $\delta_i$ denotes the distance between neighboring sample points.

\subsection{Diffusion Priors for 360$\degree{}$ Reconstruction}
\label{sec:DP}

We incorporate Zero-1-to-3~\cite{liu2023zero1to3}, a view-conditioned latent diffusion model, and distill knowledge from it to complete the unobserved region using SDS~\cite{poole2022dreamfusion}:
\begin{equation}
    \nabla_\theta\gL_{\mathrm{S}}=\sE_{i, \epsilon}\left[w(i)\left(\epsilon_\psi\left(\tilde{\rvz}_i ; \rmI_r, \Delta \rmP, i\right)-\epsilon\right) \frac{\partial \rmI_v}{\partial \theta}\right],
    \label{eq:sds}
\end{equation}

\noindent
where $\rmI_v$ is the rendered image from the sampled virtual view, which has a relative pose $\Delta \rmP$ with respect to the real image observation $\rmI_r$ at the reference view. 
Here the relative pose $\Delta \rmP = [\Delta r, \Delta \omega, \Delta \phi] = [r_v - r_r, \omega_v - \omega_r, \varphi_v - \varphi_r]$ is parameterized in polar-coordinate. 
Conditioned on the reference view $\rmI_r$, the relative pose $\Delta \rmP$, and the sampled time-step $i \sim \mathcal{U}(0, 1)$, the de-noising U-Net $\epsilon_\psi(\cdot)$ applies a de-noising step on the noisy latent $\tilde{\rvz}_i$ of the rendered view $\rmI_v$. 
$\theta$ is the learn-able parameters of our dynamic scene model. 
$w(i)$ is a time-dependent weighting term. 

\noindent
\textbf{Temporal View-dependent SDS.} As our input is a monocular video of a dynamic object, we condition the diffusion model on different frames so as to capture the motion of the target object.
We select a keyframe for every $k$ frame as the conditioning reference views of the diffusion prior. 
For any given rendered virtual view $\rmI_v^t$ at frame $t$, we choose its nearest keyframe $t_k$ as the reference view condition and modify the Eq.~\ref{eq:sds} as:
\begin{equation}
    \nabla_\theta\gL_{\mathrm{S}}^t=\sE_{i, \epsilon}\left[\hat{w}(i)\left(\epsilon_\psi\left(\tilde{\rvz}_i^t ; \rmI_r^{t_k}, \Delta \rmP^t, i\right)-\epsilon\right) \frac{\partial \rmI_v^t}{\partial \theta}\right].
    \label{eq:sds_t}
\end{equation}

\noindent
The latent vectors of each keyframe for conditioning are pre-computed to save the computational cost. 
%
%
To mitigate the negative effects of view-inconsistency stemming from the SDS loss, inspired by \cite{stabledreamfusion2023}, we incorporate a view-dependant modulation term to the weighting term $\hat{w}(i) = w(i)(\exp(|\Delta \rmP^t_{(1:2)}|)-1)$. This reduces the gradient for virtual views that have larger view differences with respect to the reference view: 
\begin{equation}
    |\Delta \rmP_{(1:2)}^t| = \arccos{\big(\frac{\rvo_r \cdot \rvo_v}{\norm{\rvo_r} \norm{\rvo_v}}\big)} / \pi, 
\end{equation}

\noindent where $\rvo_r$ and $\rvo_v$ are the Cartesian coordinates of the camera frame origins of the reference and virtual views. 

\noindent
\textbf{Implementation Details.} Following Dreamfusion~\cite{poole2022dreamfusion}, we use the shading model for rendering virtual views and randomly replace albedo with white color to produce textureless rendering. 
A random background is used for the rendered image from a sampled virtual view. 
In order to ensure the convergence of the deformation field, we perform SDS every $j$ iteration. 
The SDS loss is generated via a reparameterization trick~\cite{stabledreamfusion2023,threestudio2023}:
\begin{equation}
    \gL_S^t=\|\mathrm{stopgrad}(\rvz_i^t - \mathrm{grad}) - \rvz_i^t\|^2,
\end{equation}

\noindent
where $\mathrm{grad}=\hat{w}(i)\left(\epsilon_\psi\left(\tilde{\rvz}_i^t ; \rmI_r^{t_k}, \Delta \rmP^t, i\right)-\epsilon\right)$. 
$\gL_S^t$ will be combined with a real view loss for optimization.

\subsection{Optimization}
\label{sec:opt}

The optimization of our scene representation can be formulated as:
\begin{equation}
    \gL = \gL_{\mathrm{real}} + \gL_{\mathrm{S}} + \gL_{\mathrm{reg}}^{\mathrm{cano}} + \gL_{\mathrm{reg}}^{\mathrm{param}} .
\end{equation}

\noindent
\textbf{Real View Supervision.} Given a sampled input frame $I_t$, we randomly sample a batch of pixels and cast rays through those pixels to perform volume rendering. The real-view supervision is formulated as:
\begin{equation}
    \gL_{\mathrm{real}} = \underbrace{\gL_{\mathrm{c}} + \gL_{\mathrm{d}} + \gL_{\mathrm{m}}}_{\text{rendering loss}} + \underbrace{\gL_{\mathrm{sdf}} + \gL_{\mathrm{surf}} + \gL_{\mathrm{smooth}}}_{\text{per-point loss}}.
\end{equation}

\noindent
We use $\ell_2$ loss for color $\gL_{\mathrm{c}}$ and depth $\gL_{\mathrm{d}}$, cross-entropy loss for mask $\gL_{\mathrm{m}}$. 
In terms of the per-point loss, following~\cite{azinovic2022neural,wang2022go,wang2023co} we set up a truncation distance $tr$ and apply pseudo-SDF loss $\gL_\mathrm{sdf}$ to the ray-points within the truncation region. 
To speed up the convergence of our model, we employ $\gL_{\mathrm{surf}}$ that directly supervises the SDF and color of the back-projected surface point using the input color and depth. 
Since Hash encoding does not have global continuity, resulting in noisy surface reconstruction, we further apply a normal smoothness term $\gL_{\mathrm{smooth}}$ to near-surface points to encourage local smoothness of the SDF gradient in observation space: 
\begin{equation}
    \gL_\mathrm{smooth} = \frac{1}{|S|} \sum_{\rvx \in S} \norm{\nabla s(\rvx) - \nabla s(\rvx + \delta{\rvx})}^2,
\end{equation}
\noindent where $S$ is obtained by uniformly sampling $n$ points from $\hat{D} - tr/2$ to $\hat{D} + tr/2$ along sampled rays. 
The perturbed point $\rvx + \delta \rvx$ is sampled on a circle centered at $\rvx$ with radius $r$ orthogonal to the gradient vector $\nabla s(\rvx)$. 
Following~\cite{li2023neuralangelo}, we compute the SDF gradient via finite difference.

\noindent
\textbf{Canonical Space Regularization.} We find that performing regularization using all points on sampled rays in observation space may lead to degenerate solutions. 
Instead, we propose to apply constraints directly to each individual slice of the  hyper-dimensional canonical space by skipping the deformation network:
\begin{equation}
    \begin{aligned}
    \rvx^\prime_{\mathrm{reg}} = \{\rvx_t, T(\phi(\rvx_t), \gV_t(t)) \}.
    \label{eq:cano_ref}
    \end{aligned}
\end{equation}

\noindent
As we sample rays from 360$\degree{}$ views, ensuring that the sampled points can span the entire space across different frames, using coordinates in Eq.~\ref{eq:cano_ref} for regularization can be viewed as performing regularization in the entire hyper-dimensional canonical space. 
The regularization loss of the canonical space can be written as:
\begin{equation} \label{eq:cano_reg_all}
    \gL^{\mathrm{cano}}_{\mathrm{reg}} = \gL_{\mathrm{ori}} + \gL_{\mathrm{normal}} + \gL_{\mathrm{eik}},
\end{equation}

\noindent
where $\gL_{\mathrm{ori}}$ is the orientation loss used in DreamFusion~\cite{poole2022dreamfusion}.
$\gL_{\mathrm{eik}}$ is the eikonal loss that enforces the SDF gradient to have a unit norm. 
$\gL_{\mathrm{normal}}$ is the normal smoothness regularization on each sampled point.

\noindent
\textbf{Parameter Space Regularization.} We also perform regularization on the parameters of our model:
\begin{equation}
    \gL^{\mathrm{param}}_{\mathrm{reg}} = \gL_{\text{code}} + \gL_{\beta}.
\end{equation}

\noindent
$\gL_{\text{code}}$ is a regularization term on deformation code to encourage temporal smoothness of object motion:
\begin{equation}
    \gL_{\text{code}} = \frac{1}{|W_t|}\sum_{k \in W_t} \norm{2\gV_t(k) - \gV_t(k - 1) - \gV_t(k + 1)}^2,
\end{equation}

\noindent
where $W_t= \{\dots, t-1, t, t+1, \dots\}$ is a local window around $t$. $\gL_{\beta} $ is $\ell_1$ loss used for minimizing $\beta$ in $\sigma_\beta(s)$. 
\section{Experiments}

\subsection{Experimental Setup}

\newcommand{\tit}[2]{\multirow{2}{*}{\begin{tabular}[c]{@{}c@{}}\tt #1 \\ \tt #2\end{tabular}}}

\begin{figure*}[htbp]
  \centering
  \newcommand{\sz}{0.58}
  \setlength{\tabcolsep}{1.5pt}
  
  \begin{tabular}{c cccc c cccc}
    \tit{Reference}{Frames}& \tit{NDR}{(Mesh)} & \tit{Ours}{(Mesh)}& \tit{NDR}{(Color)}& \tit{Ours}{(Color)}&\tit{Reference}{Frames}& \tit{NDR}{(Mesh)}& \tit{Ours}{(Mesh)}& \tit{NDR}{(Color)} &\tit{Ours}{(Color)}\\
    &&&&&&&&&\\
    \makecell{\includegraphics[height=\sz\linewidth]{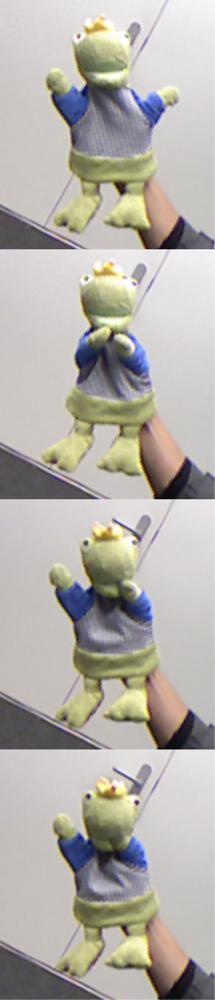}} &
    \makecell{\includegraphics[height=\sz\linewidth]{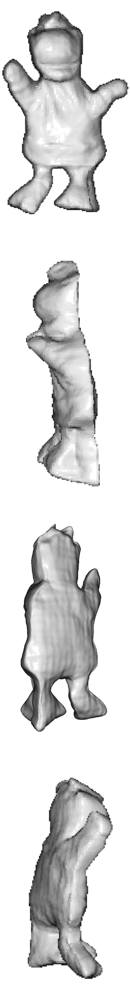}} &
    \makecell{\includegraphics[height=\sz\linewidth]{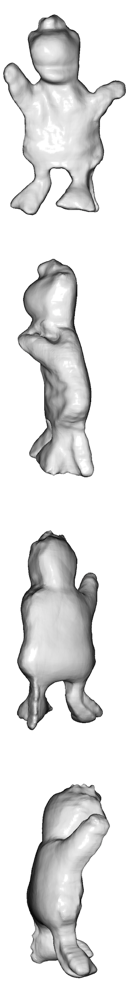}} &
    \makecell{\includegraphics[height=\sz\linewidth]{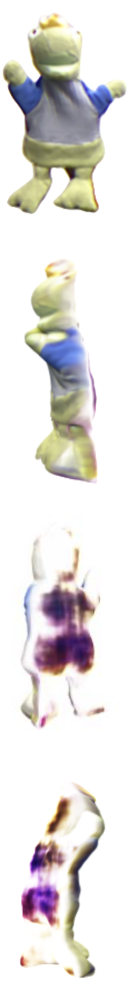}} &
    \makecell{\includegraphics[height=\sz\linewidth]{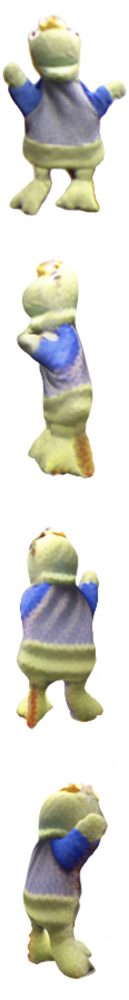}} &

    \makecell{\includegraphics[height=\sz\linewidth]{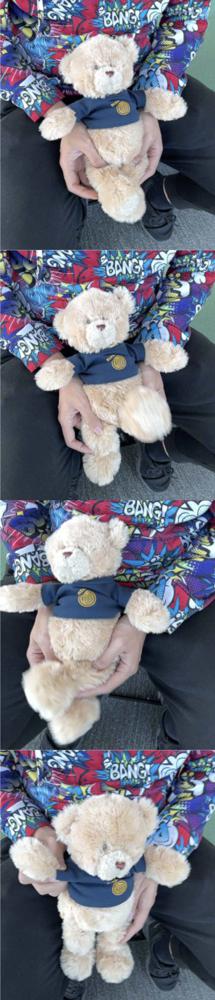}} &
    \makecell{\includegraphics[height=\sz\linewidth]{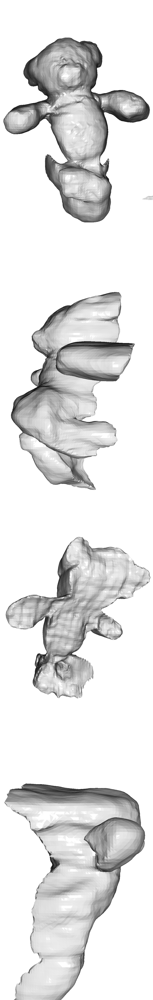}} &
    \makecell{\includegraphics[height=\sz\linewidth]{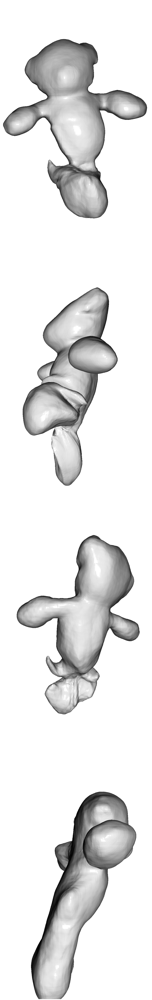}} &
    \makecell{\includegraphics[height=\sz\linewidth]{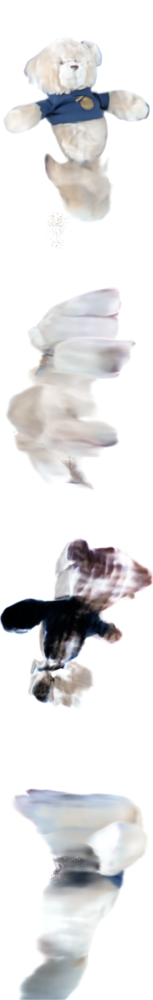}} &
    \makecell{\includegraphics[height=\sz\linewidth]{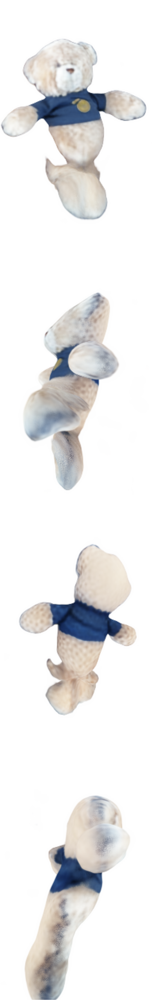}}\\


    \makecell{\includegraphics[height=\sz\linewidth]{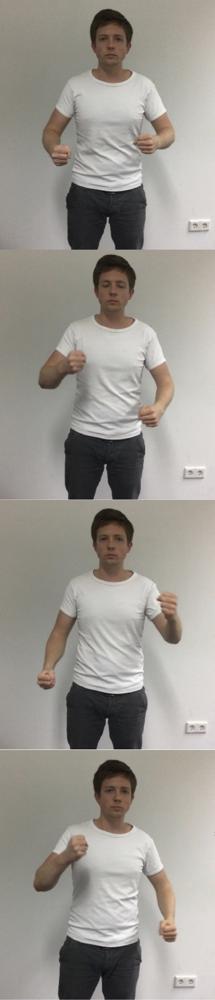}} &
    \makecell{\includegraphics[height=\sz\linewidth]{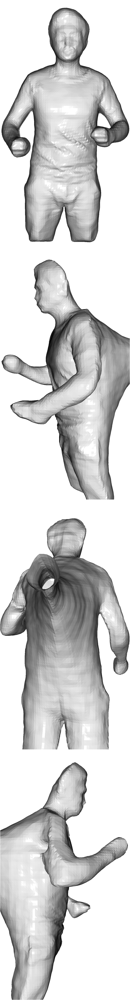}} &
    \makecell{\includegraphics[height=\sz\linewidth]{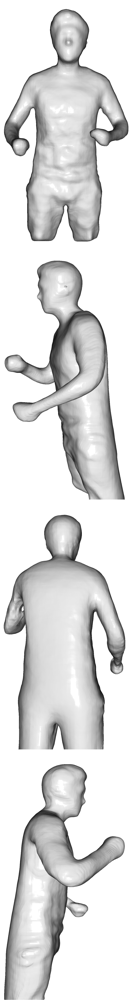}} &
    \makecell{\includegraphics[height=\sz\linewidth]{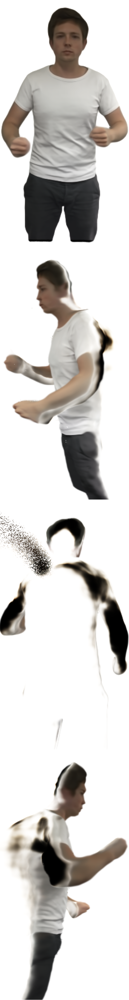}} &
    \makecell{\includegraphics[height=\sz\linewidth]{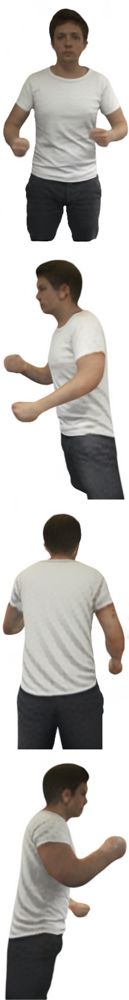}}&

    \makecell{\includegraphics[height=\sz\linewidth]{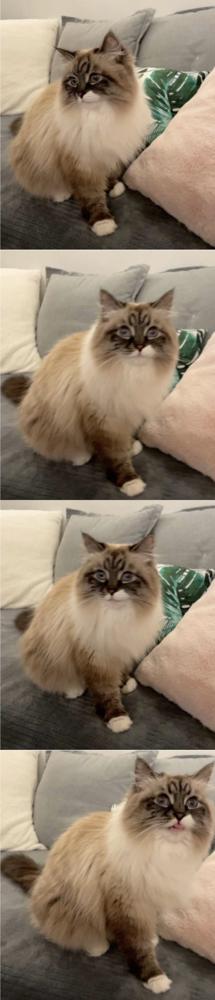}} &
    \makecell{\includegraphics[height=\sz\linewidth]{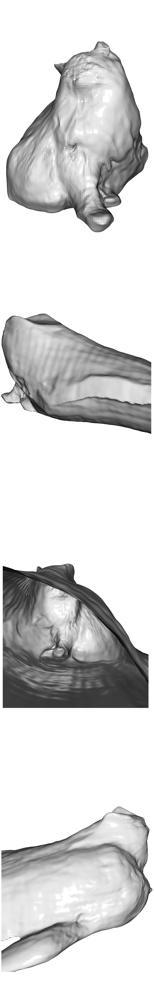}} &
    \makecell{\includegraphics[height=\sz\linewidth]{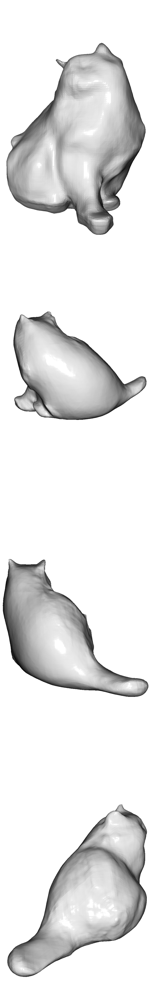}} &
    \makecell{\includegraphics[height=\sz\linewidth]{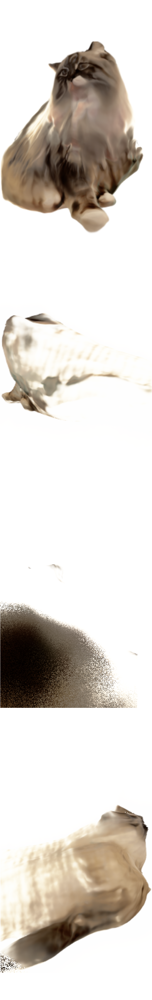}} &
    \makecell{\includegraphics[height=\sz\linewidth]{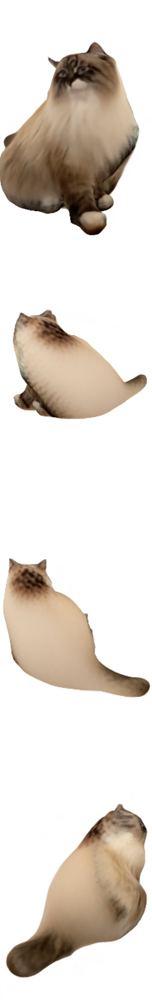}}
  \end{tabular}
  \vspace{-1mm}
  \caption{Real-world dataset reconstruction results (From left to right, top to bottom: \textsc{Frog}, \textsc{Teddy}, \textsc{Human2}, and \textsc{Mochi}). NDR~\cite{Cai2022NDR} achieves high-quality surface reconstruction in the observed region but fails to produce photo-realistic completion, resulting in spurious surfaces in unobserved regions. In contrast, our method can produce high-quality 360\degree{} surface reconstruction.}
  \label{fig:qual}
\end{figure*}
\begin{table*}[tp]
  \centering
  \footnotesize
  \setlength{\tabcolsep}{0.3em}
    \begin{tabularx}{0.95\textwidth}{c l >{\centering\arraybackslash}X >{\centering\arraybackslash}X >{\centering\arraybackslash}X >{\centering\arraybackslash}X >{\centering\arraybackslash}X >{\centering\arraybackslash}X >{\centering\arraybackslash}X >{\centering\arraybackslash}X >{\centering\arraybackslash}X >{\centering\arraybackslash}X} 
      \toprule

           \multirow{2}{*}{ Method} & \multicolumn{1}{c}{\multirow{2}{*}{Metric}} & \multicolumn{3}{c}{KillingFusion~\cite{slavcheva2017killingfusion}} & \multicolumn{3}{c}{DeepDeform~\cite{bozic2020deepdeform}} & \multicolumn{3}{c}{iPhone~\cite{gao2022monocular}}  &  \multirow{2}{*}{\makecell{\textbf{Avg.}}} \\

      \cmidrule(lr){3-5} \cmidrule(lr){6-8} \cmidrule(lr){9-11}
         & & \tt{Frog} & \tt{Duck} & \tt{Snoopy} & \tt{Human1} & \tt{Human2} & \tt{Human3} & \tt{Teddy} & \tt{Mochi} & \tt{Haru} & \\
        \midrule
        \multirow{3}{*}{\makecell{\textbf{NDR}}}   
          & {\bf Acc.} [cm] $\downarrow$
          & 1.25 & 1.34 & 1.32 & 0.74 & 0.83 & 0.83 & 6.12 & 2.19 & 2.73 & 1.92 \\
          & {\bf Comp.} [cm] $\downarrow$
          & 0.99 & 1.15 & 1.05 & 0.63 & 0.68 & 0.65 & 1.91 & 1.06 & 1.73 & 1.09 \\
          & {\bf Clip sim.} $\uparrow$
          & 82.86 & 82.52 & 80.66 & 82.22 & 83.15 & 79.24 & 75.83 & 74.12 & 77.19 & 79.75 \\
      \midrule
      \multirow{3}{*}{\makecell{\textbf{Ours}}}   
          & {\bf Acc.} [cm] $\downarrow$
          & \bf 0.77 & \bf 0.88 & \bf 1.05 & \bf 0.57 & \bf 0.61 & \bf 0.78  & \bf 0.99 & \bf 0.77 & \bf 1.51 & \bf 0.88 \\
          & {\bf Comp.} [cm] $\downarrow$
          & \bf 0.67 & \bf 0.81 & \bf 0.92 & \bf 0.49 & \bf 0.56 & \bf 0.61 & \bf 0.93 & \bf 0.60 & \bf 1.47 & \bf 0.78 \\
          & {\bf Clip sim.} $\uparrow$
          & \bf 90.24 & \bf 93.15 & \bf 90.82 & \bf 86.41 & \bf 88.75 & \bf 82.67 & \bf 80.50 & \bf 81.57 & \bf 86.87 & \bf 86.77\\
    \bottomrule
    
    \end{tabularx}%
    \vspace{-5pt}

    \caption{Per-scene quantitative results on real world datasets, including KillingFusion~\cite{slavcheva2017killingfusion}, DeepDeform~\cite{bozic2020deepdeform}, and iPhone~\cite{gao2022monocular} dataset. The 3D metrics are obtained by comparing the back-projected depth and our culled mesh of each frame. }
    \vspace{-10pt}
    \label{tab:quantitative}
\end{table*}

\begin{table}[tp]
  \centering
  \footnotesize
  \setlength{\tabcolsep}{0.3em}
    \begin{tabularx}{0.95\columnwidth}{c l >{\centering\arraybackslash}X >{\centering\arraybackslash}X >{\centering\arraybackslash}X >{\centering\arraybackslash}X >{\centering\arraybackslash}X} 
      \toprule

           \multirow{2}{*}{ Method} & \multicolumn{1}{c}{\multirow{2}{*}{Metric}} & \multicolumn{2}{c}{AMA~\cite{vlasic2008articulated}} & \multicolumn{2}{c}{BANMo~\cite{yang2022banmo}} &  \multirow{2}{*}{\makecell{Avg.}} \\
           \cmidrule(lr){3-4} \cmidrule(lr){5-6}
           & & Samba & Swing &Eagle1 & Eagle2 & \\

        \midrule

            \multirow{3}{*}{\makecell{\textbf{NDR}}}   
          & {\bf Acc.} [cm] $\downarrow$
          & 2.32 & 2.70 & 6.51 & 15.42 & 6.73\\
          & {\bf Comp.} [cm] $\downarrow$
          & 1.94 & 2.17 & 2.58 & 3.70 & 2.59\\
          & {\bf Clip sim.} $\uparrow$
          & 86.72 & 88.03 & 86.23 & 81.64 & 85.65\\

        \midrule

            \multirow{3}{*}{\makecell{\textbf{Ours}}}   
          & {\bf Acc.} [cm] $\downarrow$
          & \bf 1.98 & \bf 1.71 & \bf 2.16 & \bf 4.43 & \bf 2.57\\
          & {\bf Comp.} [cm] $\downarrow$
          & \bf 1.88 & \bf 1.83 & \bf 2.07 & \bf 3.67 & \bf 2.36\\
          & {\bf Clip sim.} $\uparrow$
          & \bf 92.48 & \bf 91.86 & \bf 90.63 & \bf 90.28& \bf 91.31\\

        \bottomrule
    \end{tabularx}%
    \vspace{-5pt}
    \caption{Per-scene quantitative results on synthetic datasets, including AMA~\cite{vlasic2008articulated} and BANMo~\cite{yang2022banmo} datasets. The 3D metrics are obtained by comparing the ground-truth mesh of each frame. }
    \vspace{-5pt}
    \label{tab:quantitative2}
\end{table}

\textbf{Datasets.} We evaluate our method on 9 real-world scenes that are from KillingFusion~\cite{slavcheva2017killingfusion}, DeepDeform~\cite{bozic2020deepdeform}, and iPhone dataset~\cite{gao2022monocular}. 
All videos in those datasets are captured with a consumer-grade RGB-D camera. 
Object masks are obtained with off-the-shelf object segmentation methods~\cite{cheng2021mivos, lin2022robust,cheng2023putting}. 
Additionally, we also evaluate our methods on 3 synthetic scenes from AMA~\cite{vlasic2008articulated} and BANMo~\cite{yang2022banmo} dataset which ground-truth meshes at all time-stamps are available. 
For AMA dataset~\cite{vlasic2008articulated}, we render depth images from the calibrated 8 views aligned with the provided GT RGB images and masks and select the front-view observations for optimisation. 
For BANMo dataset~\cite{yang2022banmo}, we follow a similar protocol except that all the RGBs, depths and masks are rendered. 
Following NDR~\cite{Cai2022NDR}, we compensate for the object's rigid motion and relative transformation in the data pre-processing stage using Robust-ICP~\cite{zhang2021fast}. 

\noindent
\textbf{Metrics.} We adopt accuracy (acc. [cm]) and completion (comp. [cm]) to evaluate the surface reconstruction quality and CLIP similarity~\cite{radford2021clip} to evaluate the realism of completion.
For 9 real-world scenes, we perform mesh culling for evaluation due to the lack of complete GT meshes. 
The CLIP similarity is computed between RGB frames and the novel view renderings generated by our optimized model. 
These renderings follow a 360\degree{} trajectory around the object, which is in motion over time.
For 3 synthetic scenes, we evaluate the complete mesh and render RGB images from the other 7 views to obtain CLIP similarity.

\noindent
\textbf{Implementation Details.} We run \papername{} on a desktop PC with an Intel Core i7-13700K CPU and NVIDIA RTX 4090 GPU with 24GB memory. 
The model takes around 2-3 hours to train and requires 10-22GB of memory depending on the the size of the input frames. 
We employ a coarse-to-fine training strategy that progressively increases the number of frequency bands of the positional encoding and hash encoding by a modulation term~\cite{park2021hypernerf, Cai2022NDR, wang2023neus2}.
Additionally, we discover that when the learning rate is small, the learning of the canonical field (with hash encoding) is significantly faster than the deformation field (with positional encoding). 
This can let the model learn a good initialization of the canonical shape. Thus, we apply a learning rate warm-up strategy. 
In the warm-up phase, we freeze the deformation field when training the novel view with $\gL_{\mathrm{S}}$ and set the timestep range of the diffusion prior to be $[0.02, 0.5]$ to ensure the model learns a good canonical shape representation. 
In the second phase, we reduce the timestep range to $[0.02, 0.2]$ to allow the model to learn accurate motion. The rendered resolution is doubled to improve the quality of the geometry and appearance completion. 


\begin{table}[tp]
\centering
\resizebox{0.95\columnwidth}{!}{
\begin{tabular}{lccc}
\toprule
           & {\bf Acc.} [cm]$\downarrow$ & {\bf Comp.} [cm]$\downarrow$ & {\bf Clip sim.} $\uparrow$ \\
\midrule
w/o diffusion prior    &         0.99 &     \bf 0.73 &         82.68          \\
w/o temporal condition &         0.92 &         0.95 &         86.01          \\
w/o angle weight       &         0.91 &         0.86 &         86.65          \\
w/o depth    &         3.24  &    4.90 &         85.35          \\
w/o canonical space  &         0.90 &     1.05 &         86.14          \\
Full model             &     \bf 0.88 &         0.78 &     \bf 86.77          \\
\bottomrule
\end{tabular}}
\vspace{-5pt}
\caption{Ablation studies on real-world datasets: a) w/o diffusion prior: optimize on real-view only; b) w/o depth: optimize with rgb only; c) w/o temporal condition: condition on first frame d) w/o angle weight: using original $w(i)$. e) w/o canonical space: model the scene as a unified 4D space-time field in world space.}
\vspace{-5pt}
\label{tab:ab_all}
\end{table}

\subsection{Evaluation}

Tab.~\ref{tab:quantitative}-\ref{tab:quantitative2} show the per-scene quantitative results on 9 real-world scenes and 4 synthetic scenes respectively.
%
%
Both results show that our model demonstrates superior accuracy and completion in surface reconstruction, outperforming NDR which exhibits spurious surface completion due to the continuity MLPs (See Teddy in Fig.~\ref{fig:qual}).
Additionally, our method achieves a significantly better CLIP similarity in comparison to NDR, thanks to the effective knowledge distillation from the diffusion prior. 

\begin{figure}[t]
  \centering
  \setlength{\tabcolsep}{0.5pt}
  \newcommand{\sz}{0.25}
  \footnotesize
  \begin{tabular}{cccc}

    \makecell{\includegraphics[height=\sz\linewidth]{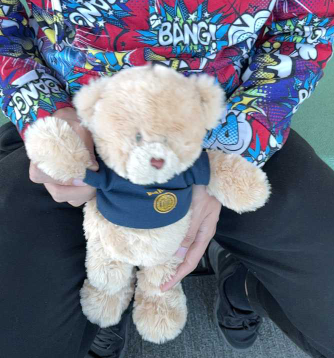}}&
    \makecell{\includegraphics[height=\sz\linewidth]{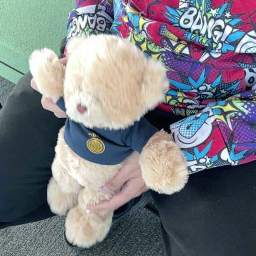}} &
    \makecell{\includegraphics[height=\sz\linewidth]{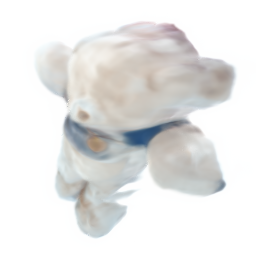}} &
    \makecell{\includegraphics[height=\sz\linewidth]{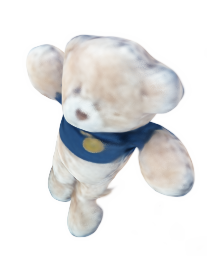}}\\

    Train view  & GT view & NDR & Ours\\
    
  \end{tabular}
  \vspace{-5pt}
  \caption{Qualitative results of novel view synthesis. Please refer to suppl. material for more quantitative \& qualitative comparisons.}
  \label{fig:nvs}
  \vspace{-5pt}
\end{figure}

Fig.~\ref{fig:qual} shows a qualitative comparison of dynamic 360\degree{} reconstruction, highlighting the ability of diffusion priors to eliminate spurious surface prediction and achieve photo-realistic completion.  
Notably, our model exhibits robustness in completing unobserved regions even in the presence of significant motion changes, as illustrated in the last row of the Teddy sequence. Fig~\ref{fig:nvs} shows a qualitative result of novel view synthesis. Training dynamic neural representations within a monocular setting typically leads to poor quality novel view synthesis in the case of large viewpoint differences. In contrast, our diffusion prior helps to alleviate this issue allowing 360\degree{} novel view synthesis. Please refer to supplementary materials for further quantitative analysis on novel view synthesis.

\subsection{Analysis}

\begin{figure}[!tbp]
  \centering
  \scriptsize
  \setlength{\tabcolsep}{1.4pt}
  \newcommand{\sz}{0.215}
  \begin{tabular}{lcccc}
    & \tt Color view 1 & \tt Color view 2 & \tt Mesh view 1 & \tt Mesh view 2 \\
     \makecell{\rotatebox{90}{w/o geo. init.}}&
    \makecell{\includegraphics[width=\sz\linewidth]{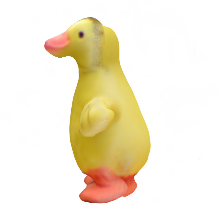}} &
    \makecell{\includegraphics[width=\sz\linewidth]{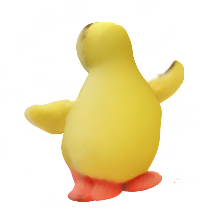}}&
    \makecell{\includegraphics[width=\sz\linewidth]{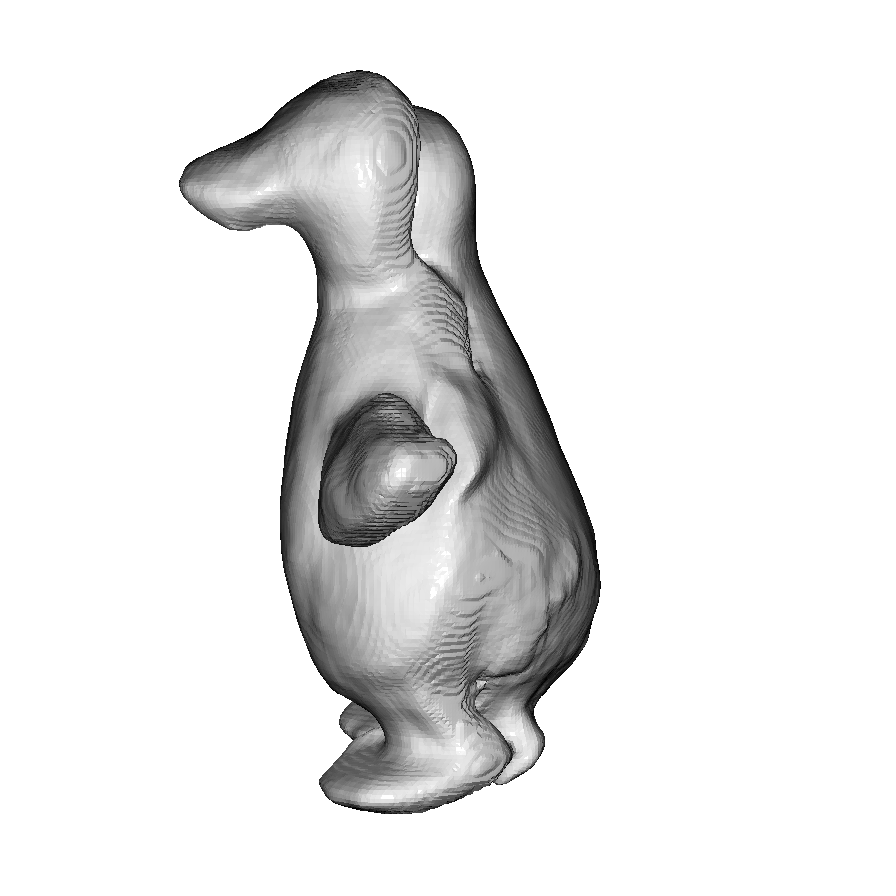}} &
    \makecell{\includegraphics[width=\sz\linewidth]{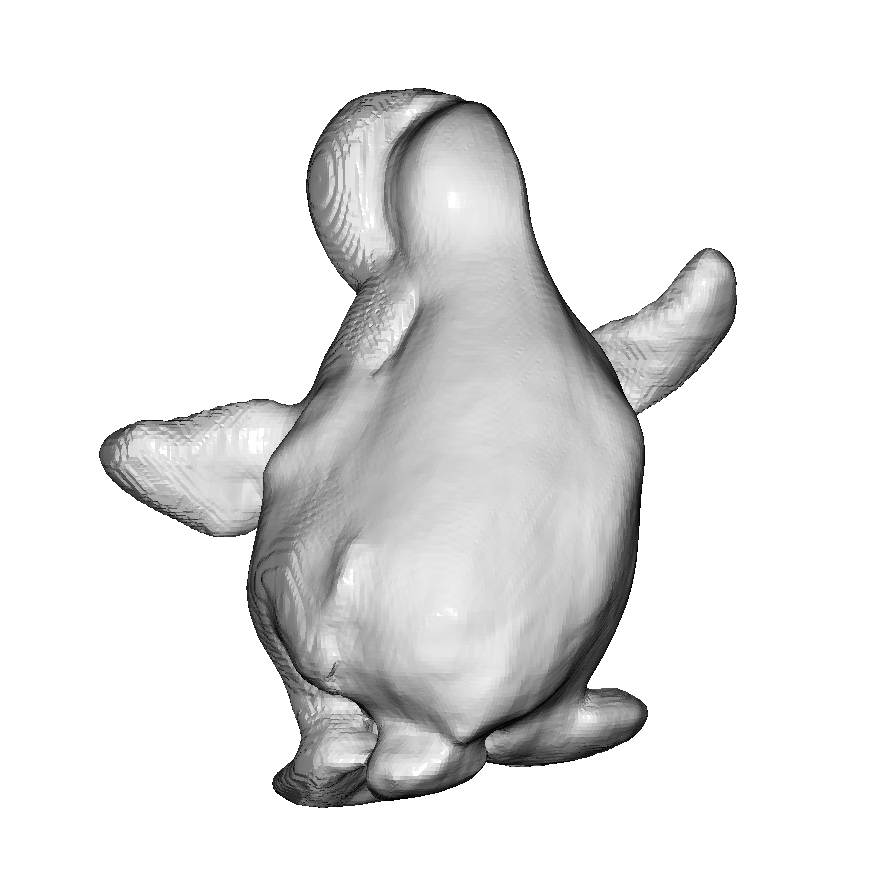}}\\
    \makecell{\rotatebox{90}{w/ geo. init.}} &
    \makecell{\includegraphics[width=\sz\linewidth]{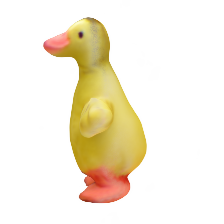}} &
    \makecell{\includegraphics[width=\sz\linewidth]{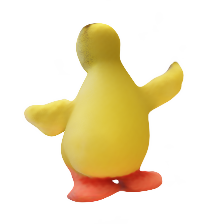}} &

    \makecell{\includegraphics[width=\sz\linewidth]{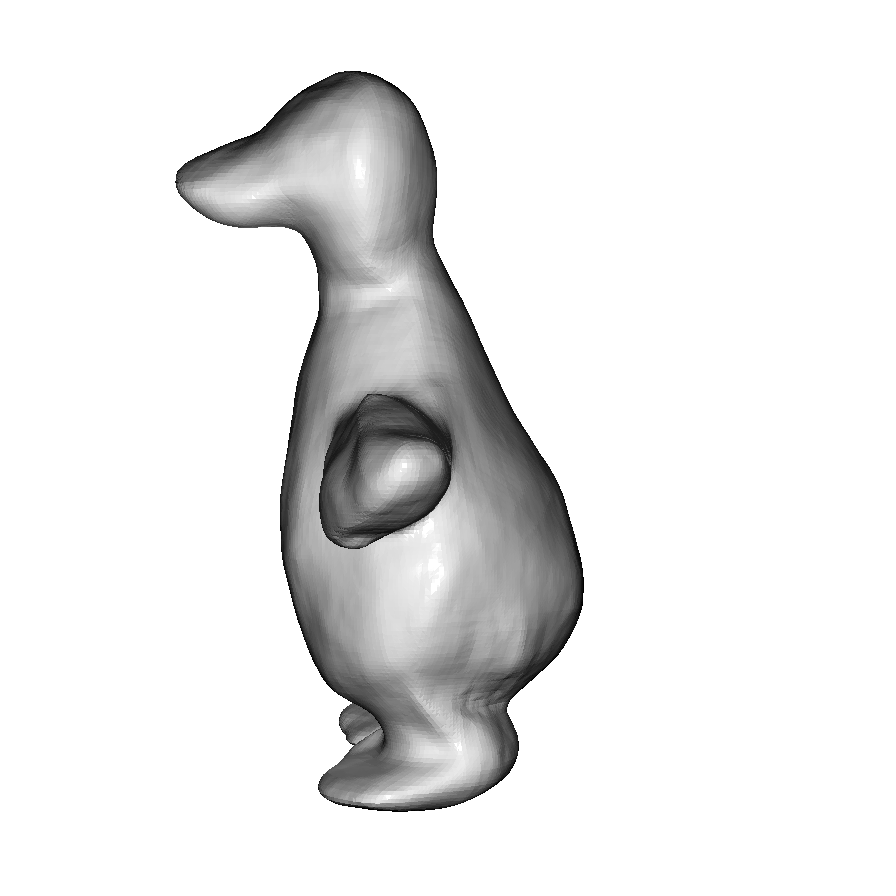}} &
    \makecell{\includegraphics[width=\sz\linewidth]{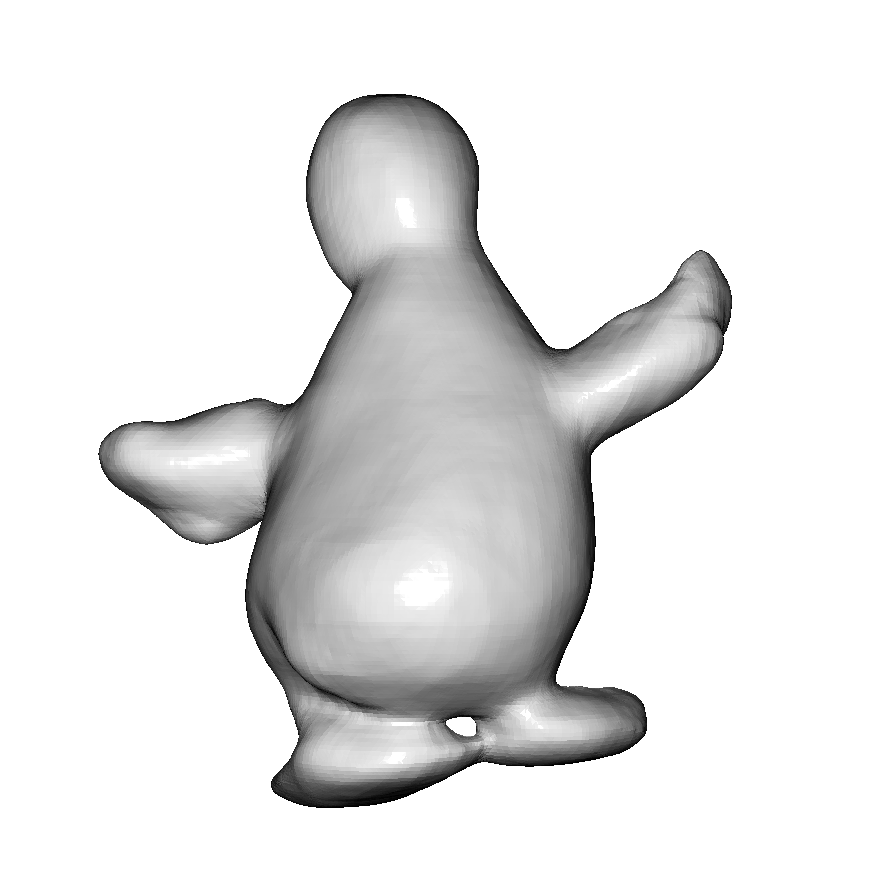}}\\
  \end{tabular}
   \vspace{-5pt}
  \caption{Ablation study on geometric initialization. Geometric initialization with fixed blob function~\cite{poole2022dreamfusion} fails here.
  }
  \label{fig:ab_init}
  \vspace{-5pt}
\end{figure}

\begin{figure}[t]
  \centering
  \footnotesize
  \setlength{\tabcolsep}{1.5pt}
  \newcommand{\sz}{0.28}
  \begin{tabular}{cccc}
    \multicolumn{2}{c}{w/o $\gL_{\mathrm{smooth}}$} & \multicolumn{2}{c}{w/ $\gL_{\mathrm{smooth}}$}\\

    \makecell{\includegraphics[height=\sz\linewidth]{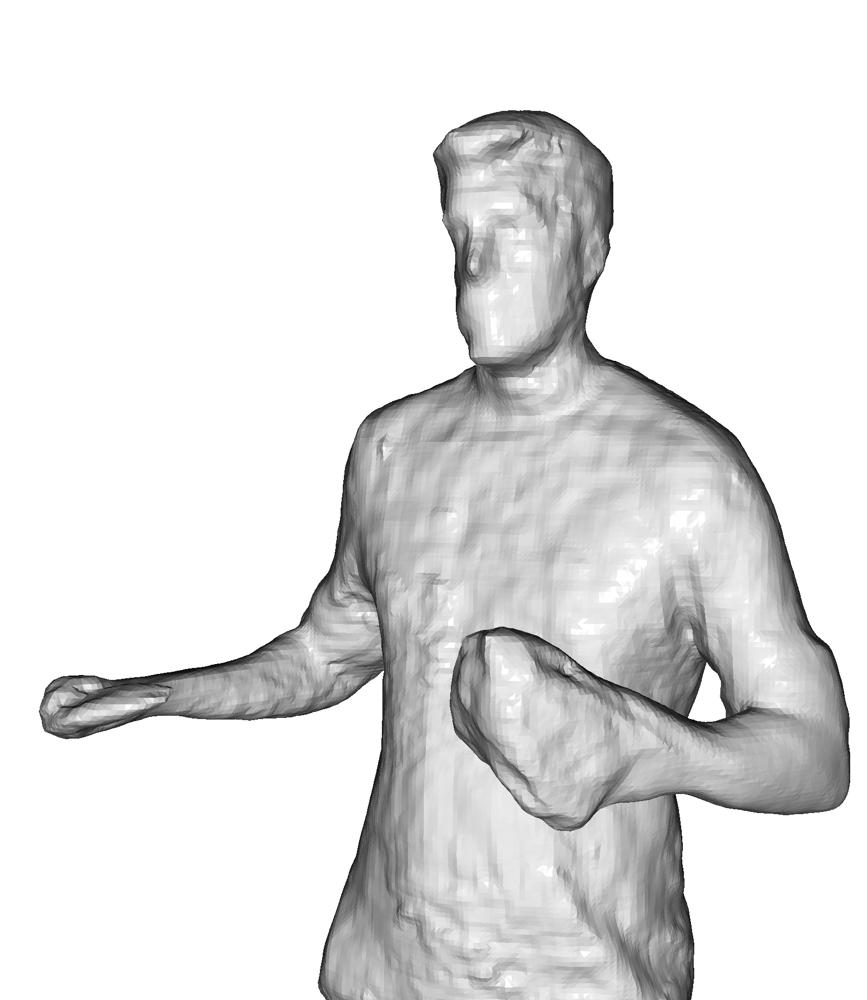}}&
    \makecell{\includegraphics[height=\sz\linewidth]{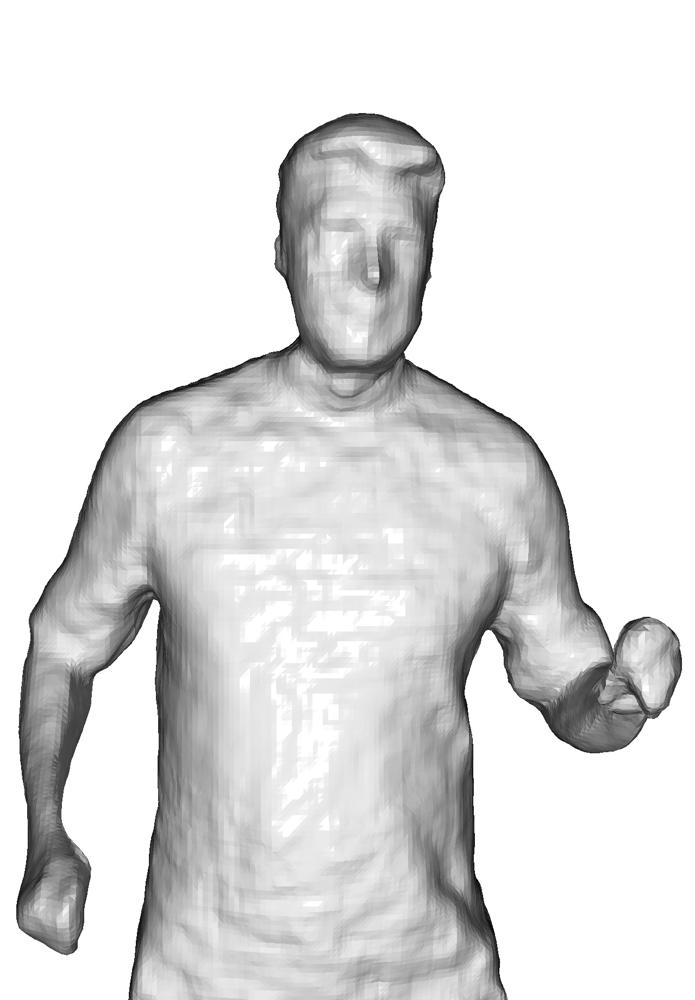}} &
    \makecell{\includegraphics[height=\sz\linewidth]{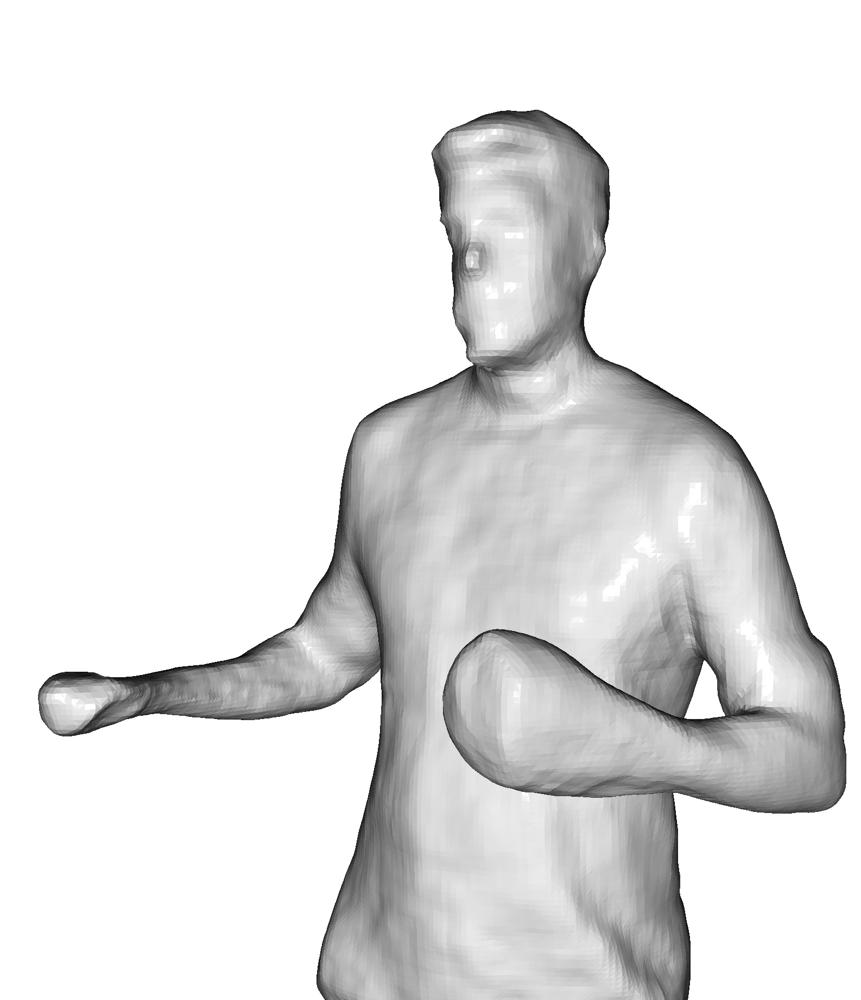}} &
    \makecell{\includegraphics[height=\sz\linewidth]{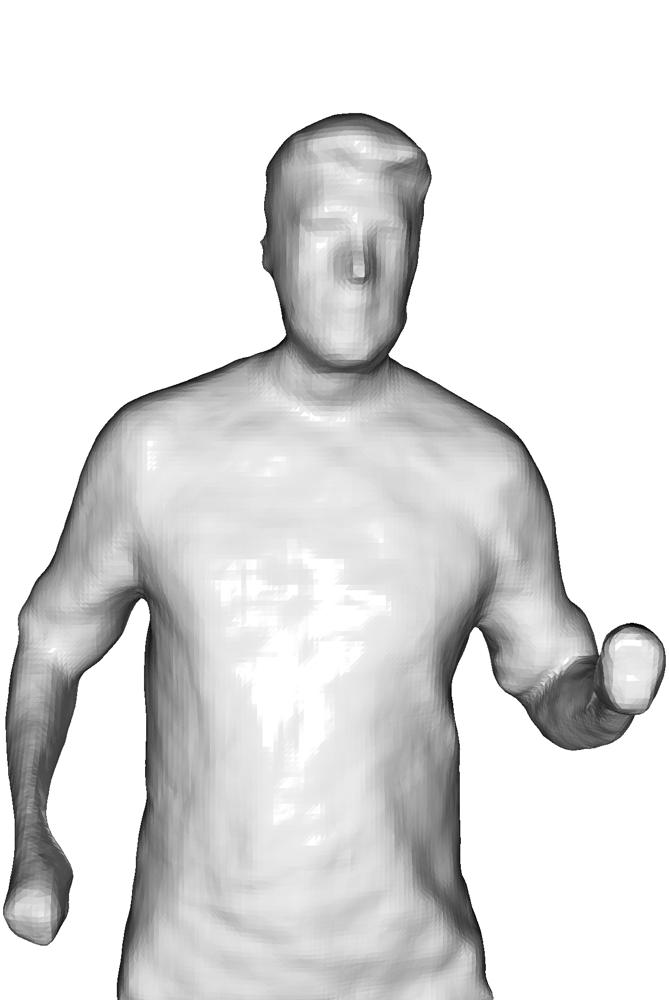}}\\
  \end{tabular}
   \vspace{-5pt}
  \caption{Ablation study on surface normal smoothness $\gL_{\mathrm{smooth}}$.}
  \label{fig:ab_smooth}
  \vspace{-5pt}
\end{figure}

\begin{figure*}[htbp]
  \centering
  \setlength{\tabcolsep}{3pt}
  \newcommand{\sz}{0.2}
  \resizebox{0.9\textwidth}{!}{
  \begin{tabular}{ccccc}
    \tt Reference& \tt Ours & \tt Ours (w/o depth)& \tt Magic123 (Coarse) & \tt Magic123 (Fine)\\
    \includegraphics[height=\sz\linewidth]{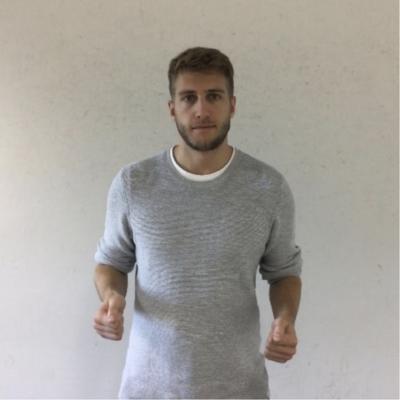} &
    \includegraphics[height=\sz\linewidth]{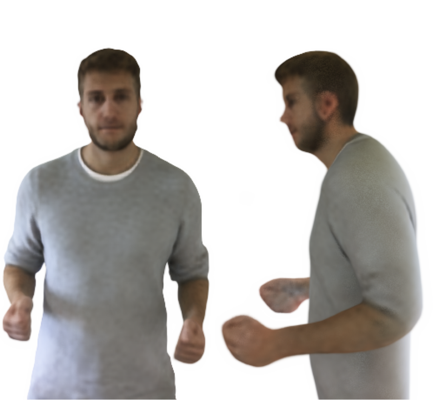} &
    \includegraphics[height=\sz\linewidth]{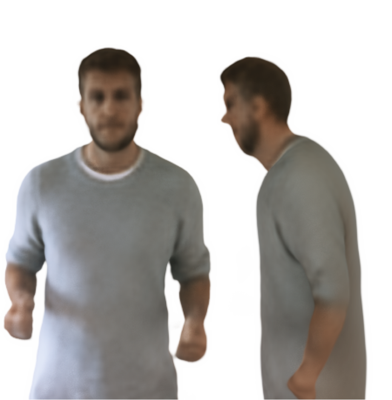} &
    \includegraphics[height=\sz\linewidth]{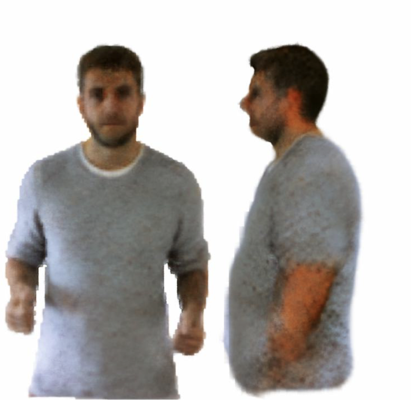} &
    \includegraphics[height=\sz\linewidth]{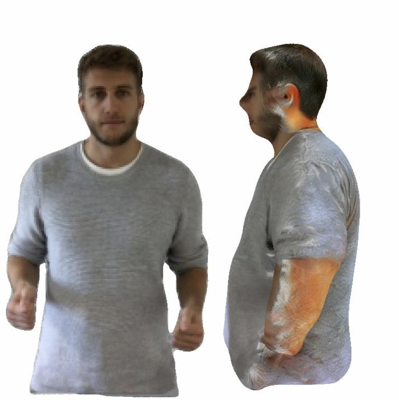} \\
    \includegraphics[height=\sz\linewidth]{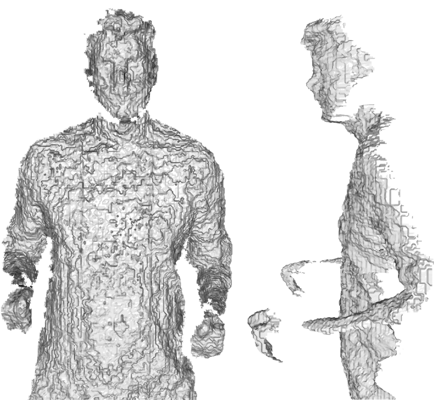} &
    \includegraphics[height=\sz\linewidth]{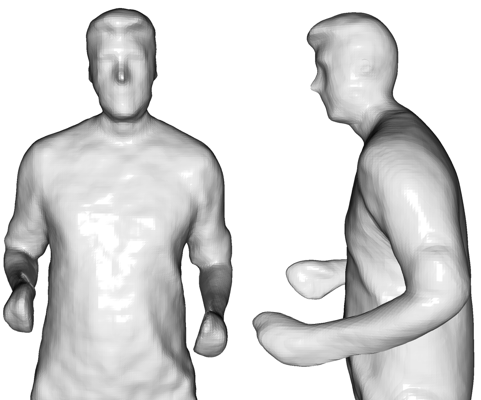} &
    \includegraphics[height=\sz\linewidth]{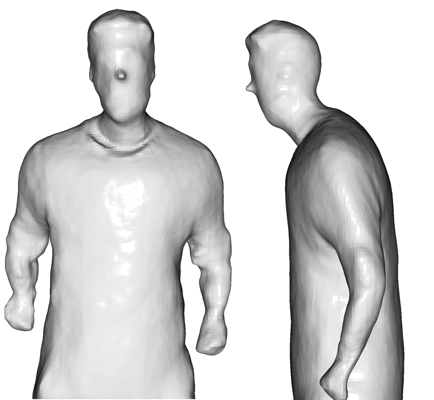} &
    \includegraphics[height=\sz\linewidth]{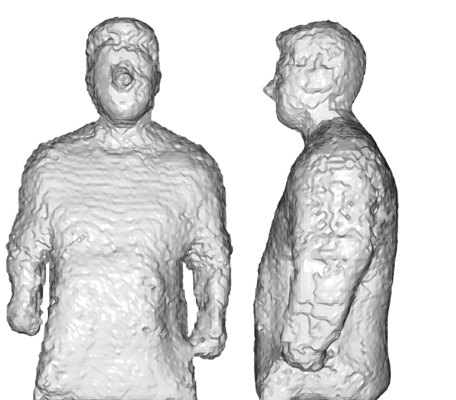} &
    \includegraphics[height=\sz\linewidth]{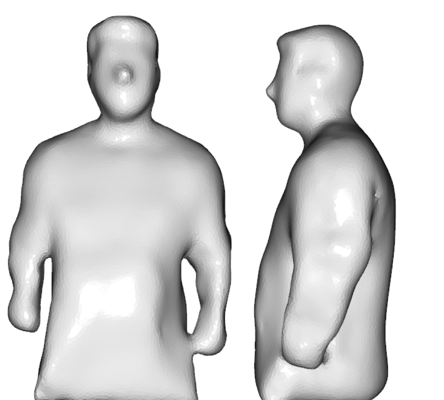}

  \end{tabular}}
  \vspace{-6pt}
  \caption{Ablation study: RGB+D vs. RGB-only.  We train Magic123~\cite{qian2023magic123} on a single frame for reference. Magic123 (Coarse) is the first stage of training using NeRF. Magic123 (fine) is the second stage of training using DMTet~\cite{shen2021dmtet} that directly optimizes surface mesh. Due to the scale ambiguity, RGB-based methods fail to capture the accurate pose of the human arm.}
  \label{fig:ab_rgb}
  \vspace{-10pt}
\end{figure*}

\begin{figure}[t]
  \centering
  \scriptsize
  \setlength{\tabcolsep}{2pt}
  \newcommand{\sz}{0.24}
  \begin{tabular}{c cc cc}
    Reference& obs. perturb.  & cano. perturb.  & w/o reg. & w/ cano reg.\\
     
    \makecell{\includegraphics[height=\sz\linewidth]{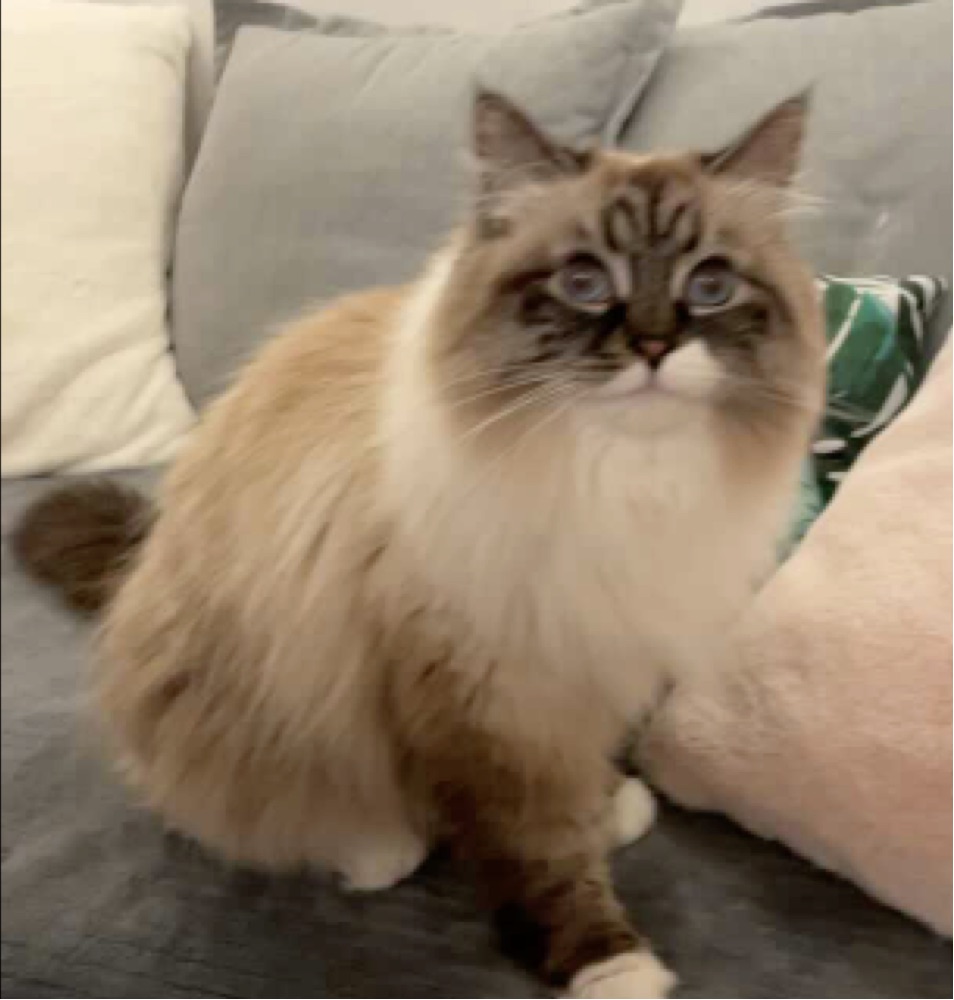}}&
    \makecell{\includegraphics[height=\sz\linewidth]{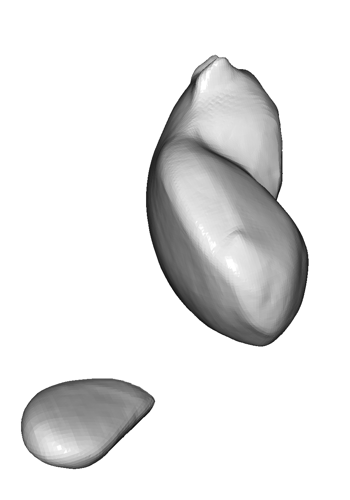}}&
    \makecell{\includegraphics[height=\sz\linewidth]{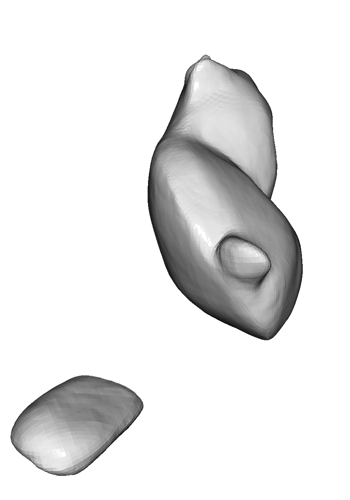}}&
    \makecell{\includegraphics[height=\sz\linewidth]{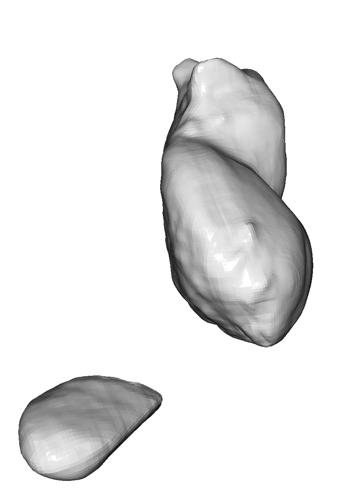}}&
    \makecell{\includegraphics[height=\sz\linewidth]{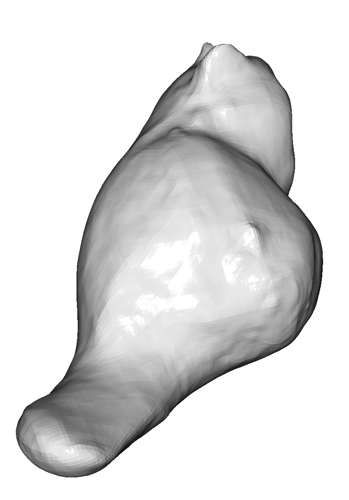}}
    \\

  \end{tabular}
  \vspace{-5pt}
  \caption{Ablation study on canonical regularization: a) obs. perturb: w/ deformation field; apply perturbation in observation space. b) cano. perturb: w/ deformation field; apply perturbation in canonical space c) w/o reg: w/o per-point normal regularization d) w/o deformation field; apply perturbation in canonical space.}
  \label{fig:ab_cano}
  \vspace{-12pt}
\end{figure}

We further analyze different components of our proposed method by conducting a set of ablation experiments on the 9 real-world scenes with quantitative results shown in Tab.~\ref{tab:ab_all} and qualitative demonstrations in Fig.~\ref{fig:ab_init}-\ref{fig:ab_cano}.

\noindent
\textbf{Temporal View-dependent SDS.} 
We start by analyzing the diffusion prior. As in Tab.~\ref{tab:ab_all}, our full model (last row) exhibits a significant improvement in CLIP similarity compared to the variant without diffusion priors (first row).
Notably, the results also highlight the diffusion priors as a regularization for surface reconstruction in the observed region, leading to the improvement in reconstruction accuracy (0.88 v.s. 0.99) by effectively eliminating spurious surfaces.

We further analyze a variant without the temporal condition (second row), where only the first frame is used as the condition for diffusion prior. 
Results show a degradation in both 3D metrics and CLIP similarity, demonstrating the importance of accurate temporal conditions for estimating the scene geometry when target object is in motion over time.
However, as opposed to using a given frame as a direct data-term supervision, the use of SDS with slightly inconsistent view conditions could still achieve realistic scene completion without drastically affecting real-view reconstruction.

Additionally, we explore a variant w/o angle weight, results in Tab.~\ref{tab:ab_all} show angle weight can effectively improve the robustness of motion capture by weighting the $\gL_{\mathrm{S}}$ based on the relative angle between real view and sampled view.

\noindent
\textbf{Geometric Initialization.} As in Fig.~\ref{fig:ab_init}, without the geometric initialization of the canonical SDF field, using SDS loss may not let the model converge to a coherent mesh. 
Note that here using fixed blob function~\cite{poole2022dreamfusion} fails to converge.

\noindent
\textbf{Normal Smoothness.} Fig.~\ref{fig:ab_smooth} shows that applying normal smoothness loss $\gL_{\mathrm{smooth}}$ to near-surface points in observation space leads to smoother reconstruction and better motion capture (e.g. hand). 

\noindent
\textbf{Effect of Depth Measurement.} As in Tab.~\ref{tab:ab_all}, although our model achieves competitive CLIP similarity without depth measurement, it fails to capture accurate motion and geometry.
We further illustrate the pitfalls of RGB-based methods in Fig.~\ref{fig:ab_rgb}. Magic123~\cite{qian2023magic123} is trained on a single frame with a monocular depth prior as a reference.
Due to the scale ambiguity, it is very challenging to recover the accurate geometry (e.g., the human arm) for RGB-based methods.

\noindent
\textbf{Canonical Space Regularization.} We also evaluate the effect of different strategies when applying our regularization losses in Eq.~\ref{eq:cano_reg_all}.
Apart from our proposed canonical space regularization (Eq.~\ref{eq:cano_ref}), we also include two variants \textit{obs. perturb.} and \textit{cano. perturb.} both of which involve the deformation network. The perturbation vector $\Delta \rvx$ is applied in the observation space and the canonical space respectively.
Results in Fig.~\ref{fig:ab_cano} reveal that both \textit{obs. perturb.} and \textit{cano. perturb.} fails to complete the tail of the cat, whereas our proposed canonical space regularization could enforce the 3D shape in hyper-canonical space to be consistent, leading to better completion of unobserved regions.

\noindent
\textbf{Effect of Representations with a Canonical Shape.} To evaluate the impact of the canonical space, we compare our full model to a canonical space-free representation (w/o canonical space). 
In the latter, the world space is directly modeled as a 4D space-time grid, which is decomposed into a quad-cube representation with 4 3D hash grids. 
Without the implicit regularization provided by a canonical space, the quad-cube demonstrates worse temporal consistency in both geometry and appearance reconstruction as in Tab.~\ref{tab:ab_all}.

\begin{figure}[t]
  \centering
  \scriptsize
  \setlength{\tabcolsep}{0.1pt}
  \newcommand{\sz}{0.24}
  \begin{tabular}{c cc}
    \multicolumn{2}{c}{Input frames}  & Reconstruction\\
    \makecell{\includegraphics[height=\sz\linewidth]{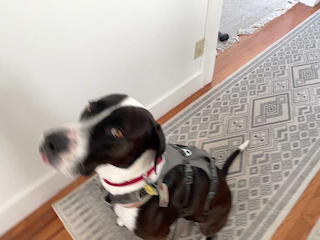}}&
    \makecell{\includegraphics[height=\sz\linewidth]{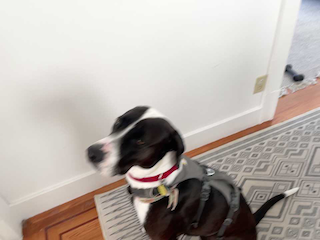}}&
    \makecell{\includegraphics[height=\sz\linewidth]{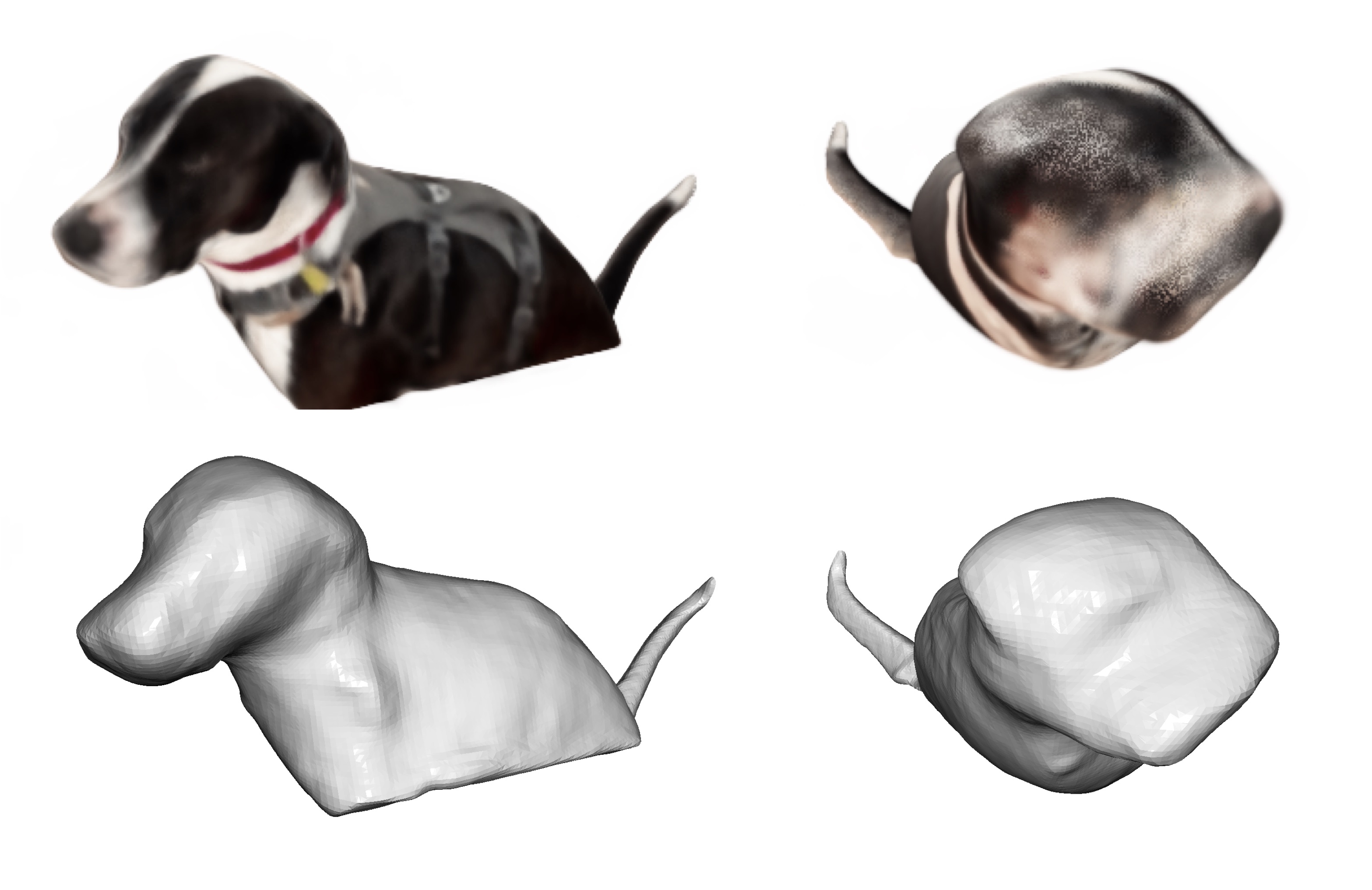}}
    \\

  \end{tabular}
  \vspace{-5pt}
  \caption{Limitations: Our method can fail in challenging situations such as incomplete views, motion blur, and challenging articulated poses of the target object.}
  \label{fig:failure}
  \vspace{-12pt}
\end{figure}

\noindent
\textbf{Limitations.} While we demonstrate photo-realistic 360\degree{} reconstruction, certain limitations are inherent in our current formulation.
Specifically, real-world captures often present challenges such as incomplete views, motion blur, and intricate articulated poses of the target object (see Fig.~\ref{fig:failure}).
Our model, which relies on the diffusion prior conditioned on pure RGB frames, may fail in these scenarios and result in incoherent reconstruction or missing geometry.
Moreover, our model cannot reconstruct complex motion in self-occluded regions due to the absence of motion priors
Introducing better diffusion priors (e.g., diffusion models that are conditioned on RGB-D images, time, etc) and motion priors can be a promising avenue for future research.

\section{Conclusion}
We presented \papername{}, a novel framework designed for dynamic 360\degree{} surface reconstruction from a casual monocular RGB-D video. 
Through the effective integration of the diffusion prior with the dynamic scene reconstruction, \papername{} goes beyond the conventional dynamic reconstruction methods by achieving both photo-realistic completion in unobserved regions and accurate motion and geometry reconstruction in the observed regions.

\noindent
\textbf{Acknowledgements.} The research presented here has been supported by a sponsored research award from Cisco Research and the UCL Centre for Doctoral Training in Foundational AI under UKRI grant number EP/S021566/1. This project made use of time on Tier 2 HPC facility JADE2, funded by EPSRC (EP/T022205/1).

{
    \small
    \bibliographystyle{ieeenat_fullname}
    \bibliography{main}
}

\end{document}


\maketitle

\section{Implementation Details}

\subsection{Data Pre-processing}
We perform a set of data pre-processing steps to the raw RGB-D sequences before training \papername{}.

\noindent
\textbf{Object Masks.}
For real-world datasets (KillingFusion, DeepDeform, iPhone), we extract foreground masks using off-the-shelf tools. Specifically, we use RVM~\cite{lin2022robust} for humans and MiVOS~\cite{cheng2021mivos} for other objects. 

\noindent
\textbf{Coordinate System.}
Following NDR~\cite{Cai2022NDR}, we subtract out the rigid motion from the target object and convert the world coordinate frame to be centered at the target object using robust ICP~\cite{zhang2021fast}. The scale of the new coordinate frame is adjusted such that the object roughly fits in a unit sphere.

\noindent
\textbf{Pseudo Observations.}
As Zero-1-to-3~\cite{liu2023zero1to3} assumes that all camera poses could be parameterised in polar coordinates (radius, polar and azimuth angles), i.e. the camera's viewing direction (z-axis) always perfectly points to the object centre (See the red cameras in Fig.~\ref{fig:pseudo-obs}). 
However, in real-world scenarios, this assumption does not hold because the camera's orientation does not depend on its translation w.r.t. the object, and thus the target object does not always appear in the middle of the image observation (See the green cameras in Fig.~\ref{fig:pseudo-obs}).
To make the diffusion prior compatible with arbitrary camera poses in practical scenarios, for every real camera pose $\rmT_{wc} = \begin{bmatrix} \rmR_{wc} & \rvt_{wc} \end{bmatrix}$ we create a pseudo camera associated with it that satisfies the polar-coordinate constraint.
The pseudo camera pose $\rmT^{\prime}_{wc} = \begin{bmatrix} \rmR^{\prime}_{wc} & \rvt^{\prime}_{wc} \end{bmatrix}$ is computed by moving the original camera center on its image plane with its orientation fixed until the camera's z-axis passes through the object center, i.e. $(0, 0, 0)$:
\begin{equation}
\rvt^{\prime}_{wc} = -\rmR_{wc}[:, 2] \cdot \rvt_{wc}, \quad
\rmR^{\prime}_{wc} = \rmR_{wc}, 
\end{equation}

\noindent
where $\rmR_{wc}[:, 2]$ denotes the last column of the camera-to-world rotation matrix. Fig.~\ref{fig:pseudo-obs} (first row) shows two examples in the snoopy and duck sequence, where real and pseudo cameras are shown in green and red respectively.
We further create pseudo-observations from those pseudo cameras by projecting the object center $(0, 0, 0)$ onto the image plane with pseudo camera poses to obtain the center pixel location for each frame and then cropping the raw image observations (RGB, depth and object mask) around the center pixel. See Fig.~\ref{fig:pseudo-obs} (second row) for a demonstration.

\noindent
\textbf{Synthetic Dataset.} For the 4 synthetic sequences (AMA-samba, AMA-swing, Eagle-1, Eagle-2) with per-frame GT meshes and multi-view real image observations, we perform the previous step for all the available camera views, but only one view is selected for optimizing our model. 

\definecolor{myred}{rgb}{1.0,0.0,0.0}
\definecolor{mygreen}{rgb}{0.0,1.0,0.0}

\begin{figure}[t]
  \centering
  \scriptsize
  \setlength{\tabcolsep}{0.1pt}
  \newcommand{\sz}{0.35}
  \begin{tabular}{l c c}
    \rotatebox[origin=c]{90}{\texttt{camera poses}} &
    \makecell{\includegraphics[height=\sz\linewidth]{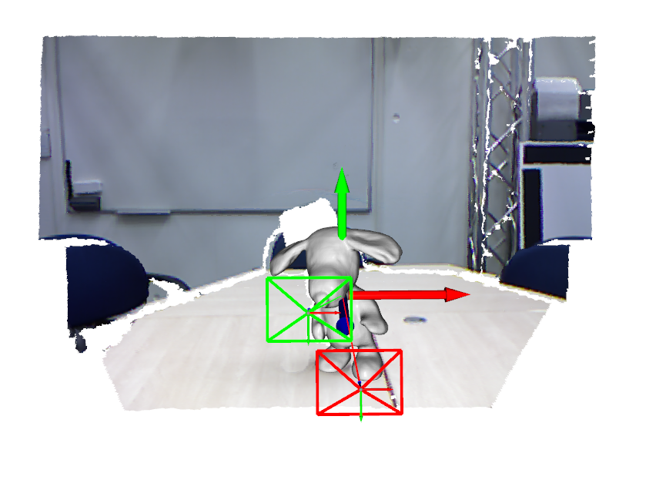}}&
    \makecell{\includegraphics[height=\sz\linewidth]{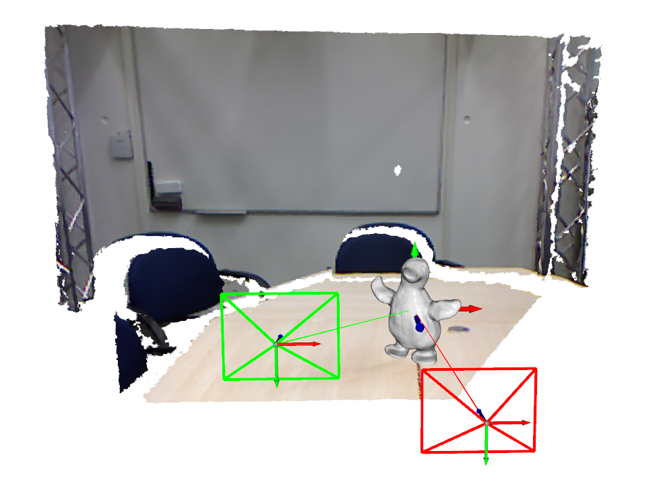}} \\
    \rotatebox[origin=c]{90}{\texttt{cropped obs.}} &
    \makecell{\includegraphics[height=\sz\linewidth]{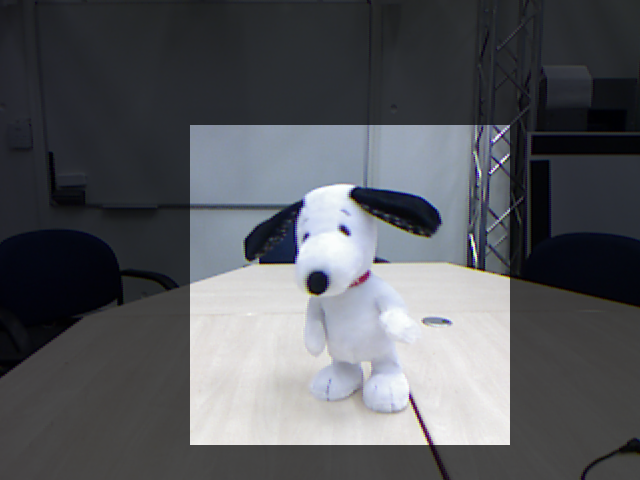}} &
    \makecell{\includegraphics[height=\sz\linewidth]{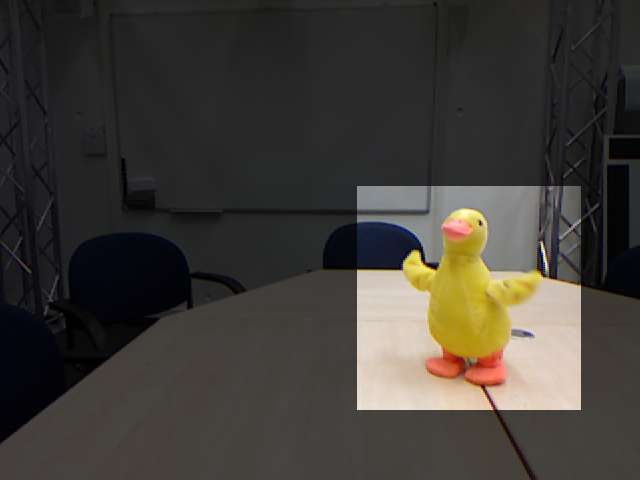}} \\
    & \texttt{snoopy} & \texttt{duck} \\
  \end{tabular}
  \caption{\textbf{Data Pre-processing.} In order to make the diffusion prior compatible with arbitrary real-world camera poses (shown in \textcolor{mygreen}{green}) in casually captured video sequences, we create pseudo cameras that directly point to object center (shown in \textcolor{myred}{red}) and obtain the pseudo-observations from the raw image.}
  \label{fig:pseudo-obs}
  \vspace{-5pt}
\end{figure}

\subsection{Hyper-parameters}

\noindent
\textbf{Deformation Field.} Our deformation field consists of 3 major components: 1) Multi-resolution deformation code $\gV_t(\cdot)$, 2) Deformation network $D(\cdot)$, and 3) Topology network $T(\cdot)$.
The multi-resolution code has 3 levels, with the resolution of $[N/8, N/4, N]$, where $N$ is the number of frames in the sequence. The feature dimension of each level is set to be $16$. 
For the deformation network and topology network, the number of the frequency band for positional encoding is set to be $6$ and $4$ respectively. The MLP used for those two networks consists of $6$ hidden layers with $128$ hidden units. The dimension of the ambient coordinate $\rvx^{\prime}_a$ predicted by the topology network is set to be $2$.

\noindent
\textbf{Canonical Field}. We represent the SDF and color of the canonical field with two Hash grids $\gV_s$ and $\gV_c$. 
Both Hash grids have $16$ levels, and the feature dimension at each level is set to $2$. 
The Hash features $\gV_s(\rvx^{\prime}_m)$ and $\gV_c(\rvx^{\prime}_m)$ are obtained via concatenating the tri-linear interpolated feature vectors at each level. 
In order to perform the geometric initialization, the Hash feature of the SDF field $\gV_s(\rvx^{\prime}_m)$ is concatenated with the 3D coordinate of the query point $\rvx^{\prime}_m$. Ideally, one can also use joint encoding strategy~\cite{wang2023co} as long as preserving 3D coordinates only and masking out the rest part.
The SDF decoder $f_{\gamma}(\cdot)$ takes in the 3D coordinate, Hash feature, and the ambient coordinate, predicts the SDF value and a $16$-D geometric feature $\rvh$.
The geometric feature and Hash feature of the color grid are then fed to the color decoder $f_{\alpha}(\cdot)$ for decoding the color values.
Both decoders are 3-layer MLPs with $64$ hidden units.

\noindent
\textbf{Optimization.} We train \papername{} with Adam~\cite{kingma2014adam} optimizer and an EMA decaying of $0.95$ for $E_{\max}=2000$ epochs.
We adopt the following scheduling strategy for the learning rate $\mu$:
\begin{equation}
\mu = 
    \begin{cases}
        \mu_1 & \text { if } E \leq 0.5E_w \\ 
        \mu_1 + \frac{2E - E_w}{E_w}(\mu_2-\mu_1)& \text { if }E\leq E_w\\
        \mu_2(\cos(\frac{E - E_w}{E_{\max} - E_w}\pi)\frac{1-k}{2} + \frac{1+k}{2})& \text { if } E > E_w\\
    \end{cases},
\end{equation}

\noindent
where $E$ is the epoch and $E_w=200$ is the number of warm-up epochs. 
A small initial learning rate $\mu_1=5e-6$ is used for better initializing the canonical field during the first phase of the warm-up stage ($E \leq 0.5E_w$). 
In the second phase ($0.5E_w < E\leq E_w$) of the warm-up stage, the learning rate is then linearly increased to $\mu_2=5e-4$.
After the warm-up stage, the learning rate is scheduled following a cosine annealing protocol. The value of $k$ is set to be $0.05$.

For each epoch, the optimization alternates between real and virtual views and the ratio of sampled virtual views and real views is set to be $0.1$. 
For the training of real view, at each iteration, we randomly sample a batch of $2048$ rays from one single frame.
For the training of the virtual view, we render the full image of the frame with down-sampled resolution. The resolution is set to be around $64\times64$ in the warm-up stage and $128\times128$ in the second stage to fit our $24$G GPU memory. 
NeRFAcc~\cite{li2023nerfacc} is used to speed up the training via efficient sampling. 
The resolution of the occupancy grid is set to be $128$, and the render step size is set to be $0.01$.

In order to achieve more robust optimization and speed up the convergence, we adopt a coarse-to-fine training strategy, where a modulation ratio term is used to control the bandwidth of the Hash grid and coordinate encoding $\lambda^b$:
\begin{equation}
    \lambda^b = \min{(0.25 + \frac{E}{E_{\max}}, 1.0)} \cdot \lambda_{\max}^b.
\end{equation}

We use Zero-1-to-3~\cite{liu2023zero1to3} as our diffusion prior. The guidance scale is set to $5.0$. The time-step range is $[0.02, 0.5]$ in the warm-up stage and $[0.02, 0.2]$ in the second stage.

\begin{figure*}[t]
  \centering
  \scriptsize
  \setlength{\tabcolsep}{1pt}
  \newcommand{\sz}{0.135}
  \begin{tabular}{l cccc ccc}
     \makecell{\rotatebox{90}{Train view}}&
    \makecell{\includegraphics[width=\sz\linewidth]{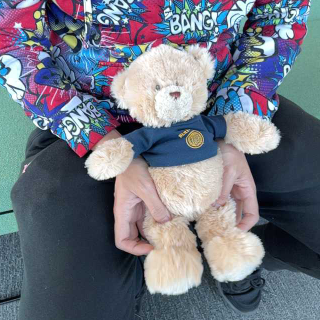}} &
    \makecell{\includegraphics[width=\sz\linewidth]{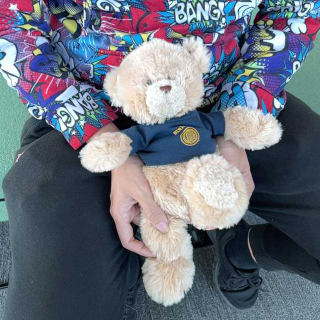}}&
    \makecell{\includegraphics[width=\sz\linewidth]{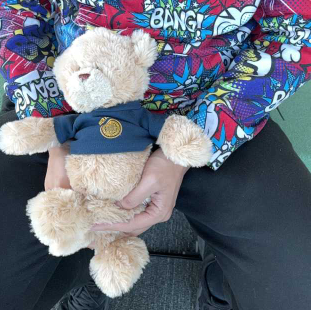}} &
    \makecell{\includegraphics[width=\sz\linewidth]{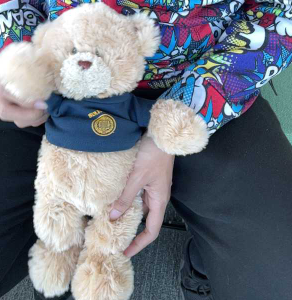}} &
    \makecell{\includegraphics[width=\sz\linewidth]{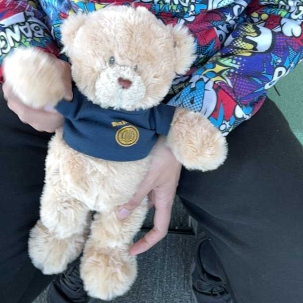}} &
    \makecell{\includegraphics[width=\sz\linewidth]{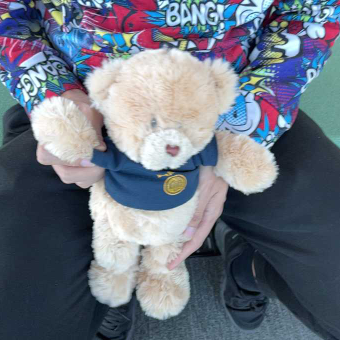}} &
    \makecell{\includegraphics[width=\sz\linewidth]{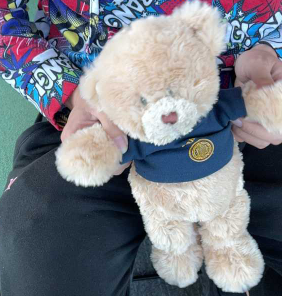}}\\
   \makecell{\rotatebox{90}{GT view}}&
    \makecell{\includegraphics[width=\sz\linewidth]{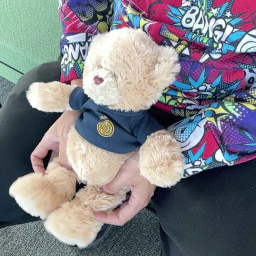}} &
    \makecell{\includegraphics[width=\sz\linewidth]{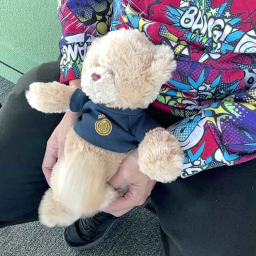}}&
    \makecell{\includegraphics[width=\sz\linewidth]{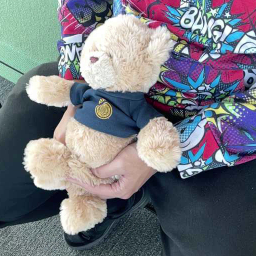}} &
    \makecell{\includegraphics[width=\sz\linewidth]{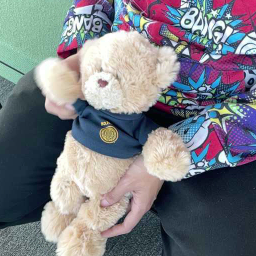}} &
    \makecell{\includegraphics[width=\sz\linewidth]{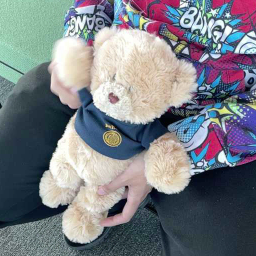}} &
    \makecell{\includegraphics[width=\sz\linewidth]{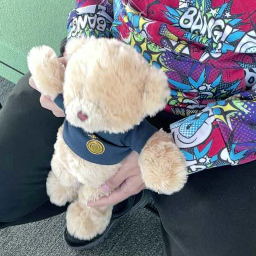}} &
    \makecell{\includegraphics[width=\sz\linewidth]{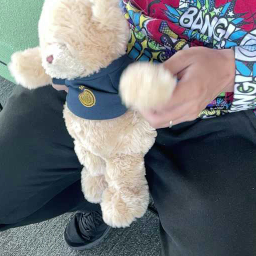}}\\

    \makecell{\rotatebox{90}{NDR}}&
    \makecell{\includegraphics[width=\sz\linewidth]{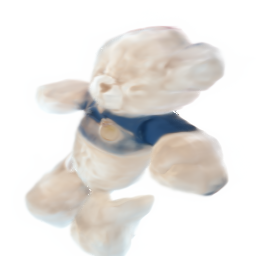}} &
    \makecell{\includegraphics[width=\sz\linewidth]{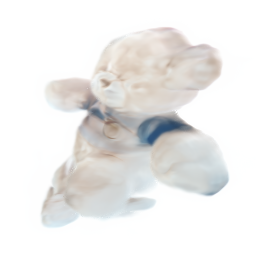}}&
    \makecell{\includegraphics[width=\sz\linewidth]{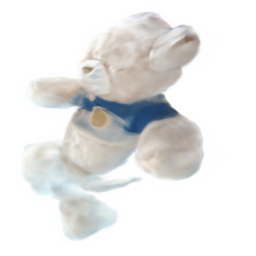}} &
    \makecell{\includegraphics[width=\sz\linewidth]{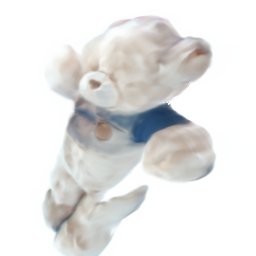}} &
    \makecell{\includegraphics[width=\sz\linewidth]{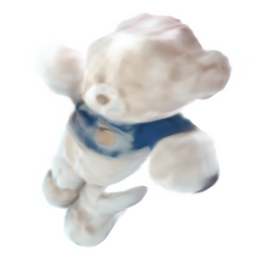}} &
    \makecell{\includegraphics[width=\sz\linewidth]{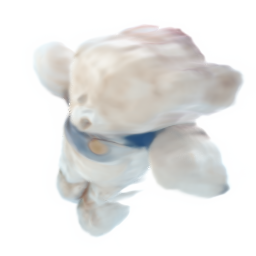}} &
    \makecell{\includegraphics[width=\sz\linewidth]{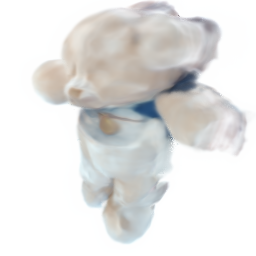}}\\
    
    \makecell{\rotatebox{90}{Ours}}&
    \makecell{\includegraphics[width=\sz\linewidth]{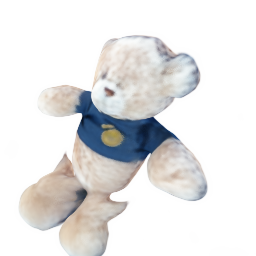}} &
    \makecell{\includegraphics[width=\sz\linewidth]{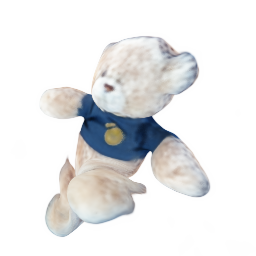}}&
    \makecell{\includegraphics[width=\sz\linewidth]{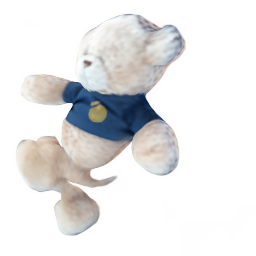}} &
    \makecell{\includegraphics[width=\sz\linewidth]{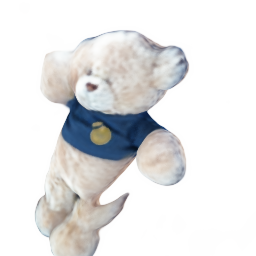}} &
    \makecell{\includegraphics[width=\sz\linewidth]{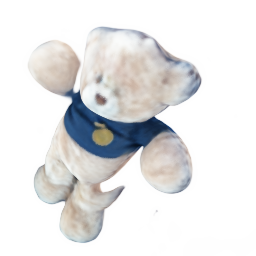}} &
    \makecell{\includegraphics[width=\sz\linewidth]{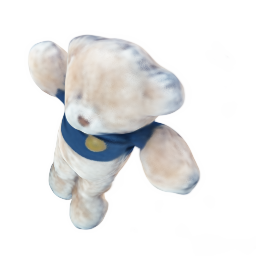}} &
    \makecell{\includegraphics[width=\sz\linewidth]{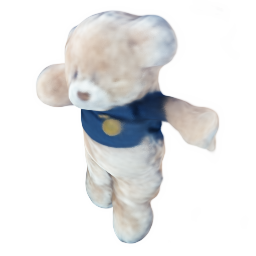}}\\
  \end{tabular}
  \caption{\textbf{Qualitative results on novel view synthesis.} We evaluate on Teddy scene which is the only sequence among all real-world scenes used in this paper that has additional GT views for evaluation. Thanks to the use of diffusion priors, our method can achieve high-quality novel synthesis given a monocular RGB-D video. 
  }
  \label{fig:supp_novel}
\end{figure*}
\section{Additional Analysis}

\subsection{Novel view synthesis}
We show additional quantitative and qualitative results in Tab~\ref{tab_r:novel} and Fig.~\ref{fig:supp_novel} respectively. For quantitative results on real-world datasets, Teddy is the only sequence that has GT reference views for evaluation. Since our training data is monocular RGB-D data, there is only one viewpoint for each timestamp. This makes novel view synthesis in this problem setting quite challenging. From the results, we could find that our method produces competitive results on novel view synthesis with consistently better perceptual quality thanks to the use of diffusion prior. Note that due to data preprocessing and the optimization of the camera pose, the non-semantic metrics, PSNR, and SSIM, may not reflect the perceptual quality. A small shift in camera poses will significantly affect those metrics, especially given the fact that we need to estimate the masked PSNR and SSIM, i.e., mPSNR and mSSIM.

\begin{table}[tp]
  \centering
  \footnotesize
  \setlength{\tabcolsep}{0.3em}
    \begin{tabularx}{0.95\columnwidth}{c l >{\centering\arraybackslash}X >{\centering\arraybackslash}X >{\centering\arraybackslash}X >{\centering\arraybackslash}X >{\centering\arraybackslash}X >{\centering\arraybackslash}X} 
      \toprule

           \multirow{2}{*}{ Method} & \multicolumn{1}{c}{\multirow{2}{*}{Metric}} & \multicolumn{2}{c}{AMA} & \multicolumn{2}{c}{BANMo} & iPhone &  \multirow{2}{*}{\makecell{Avg.}} \\
           \cmidrule(lr){3-4} \cmidrule(lr){5-6} \cmidrule(lr){7-7}
           & & Samba & Swing &eagle1 & eagle2 & Teddy & \\

        \midrule

            \multirow{3}{*}{\makecell{\textbf{NDR}}}   
          & {\bf mPSNR.}$\uparrow$
          & 7.97 & 9.89 & 14.29 & 14.80 & \bf 9.26 & 11.24\\
          & {\bf mSSIM}$\uparrow$
          & 0.326 & 0.397 & 0.241 & 0.263 &\bf 0.254 & 0.296\\
          & {\bf mLPIPS} $\downarrow$
          & 0.457 & 0.463 & 0.514 & 0.497 & 0.442 & 0.475\\

        \midrule

            \multirow{3}{*}{\makecell{\textbf{Ours}}}   
          & {\bf mPSNR.}$\uparrow$
          &\bf 10.73 &\bf 11.35 &\bf 15.37 &\bf 16.93 &9.02 & \bf 12.68\\
          & {\bf mSSIM}$\uparrow$
          &\bf 0.493 &\bf 0.510 &\bf 0.269 &\bf 0.319 &0.239 & \bf 0.366\\
          & {\bf mLPIPS} $\downarrow$
          &\bf 0.328 &\bf 0.354 &\bf 0.507 &\bf 0.447 &\bf 0.360 &\bf 0.399\\

        \bottomrule
    \end{tabularx}%
    \vspace{-5pt}
    \caption{\textbf{Per-scene quantitative results on novel view synthesis.} For synthetic datasets, we compare 8 GT views that span 360 degrees. For real-world dataset, we compare 2 GT views provided. Note that Teddy is the only sequence among all real-world scenes used in this paper that has additional GT views for evaluation. Due to the data preprocessing and the optimization of the camera pose, the non-semantic metrics here may not accurately reflect the actual performance.}
    \vspace{-5pt}
    \label{tab_r:novel}
\end{table}

\newcommand{\tit}[2]{\multirow{2}{*}{\begin{tabular}[c]{@{}c@{}}\tt #1 \\ \tt #2\end{tabular}}}

\begin{figure*}[t]
  \centering
  \scriptsize
  \setlength{\tabcolsep}{3pt}
  \newcommand{\sz}{0.35}
  \begin{tabular}{ccccc}
    \tit{Reference}{Frames} & \tt Point-E~\cite{nichol2022point} & \tt One-2-3-45~\cite{liu2023one} & \tit{Zero-1-to-3$^\star$}{(Coarse)~\cite{liu2023zero1to3}} & \tit{Zero-1-to-3$^\star$}{(Fine)~\cite{liu2023zero1to3}}\\
    & & & & \\
    \includegraphics[height=\sz\linewidth]{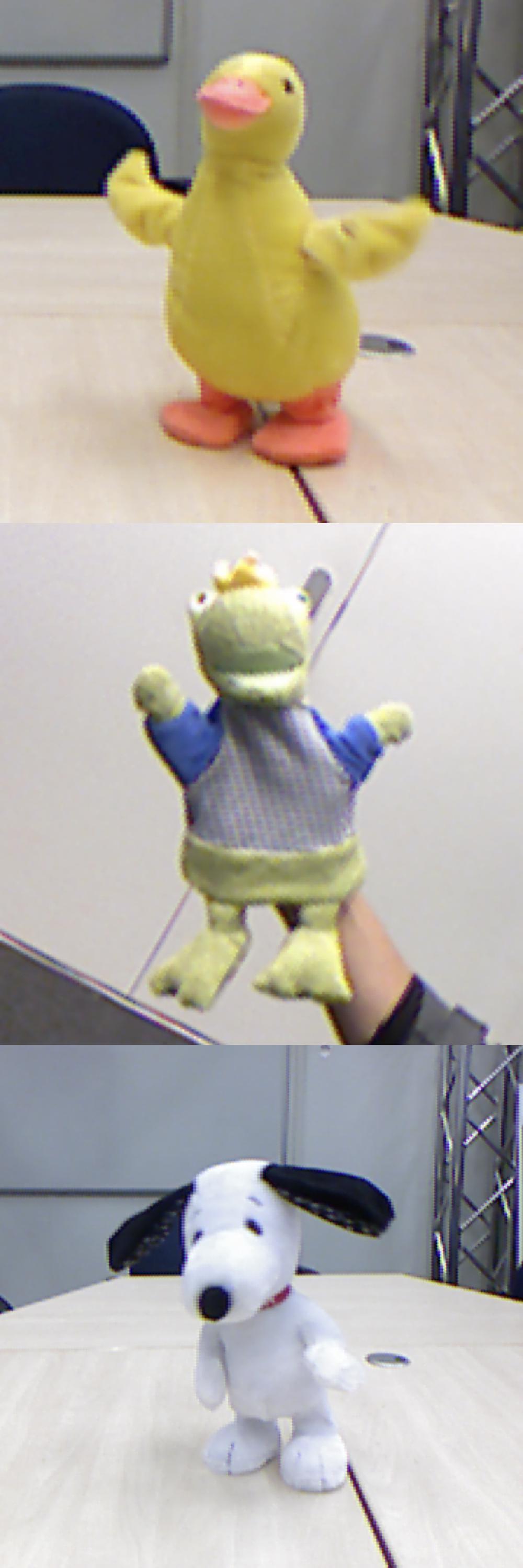} &
    \includegraphics[height=\sz\linewidth]{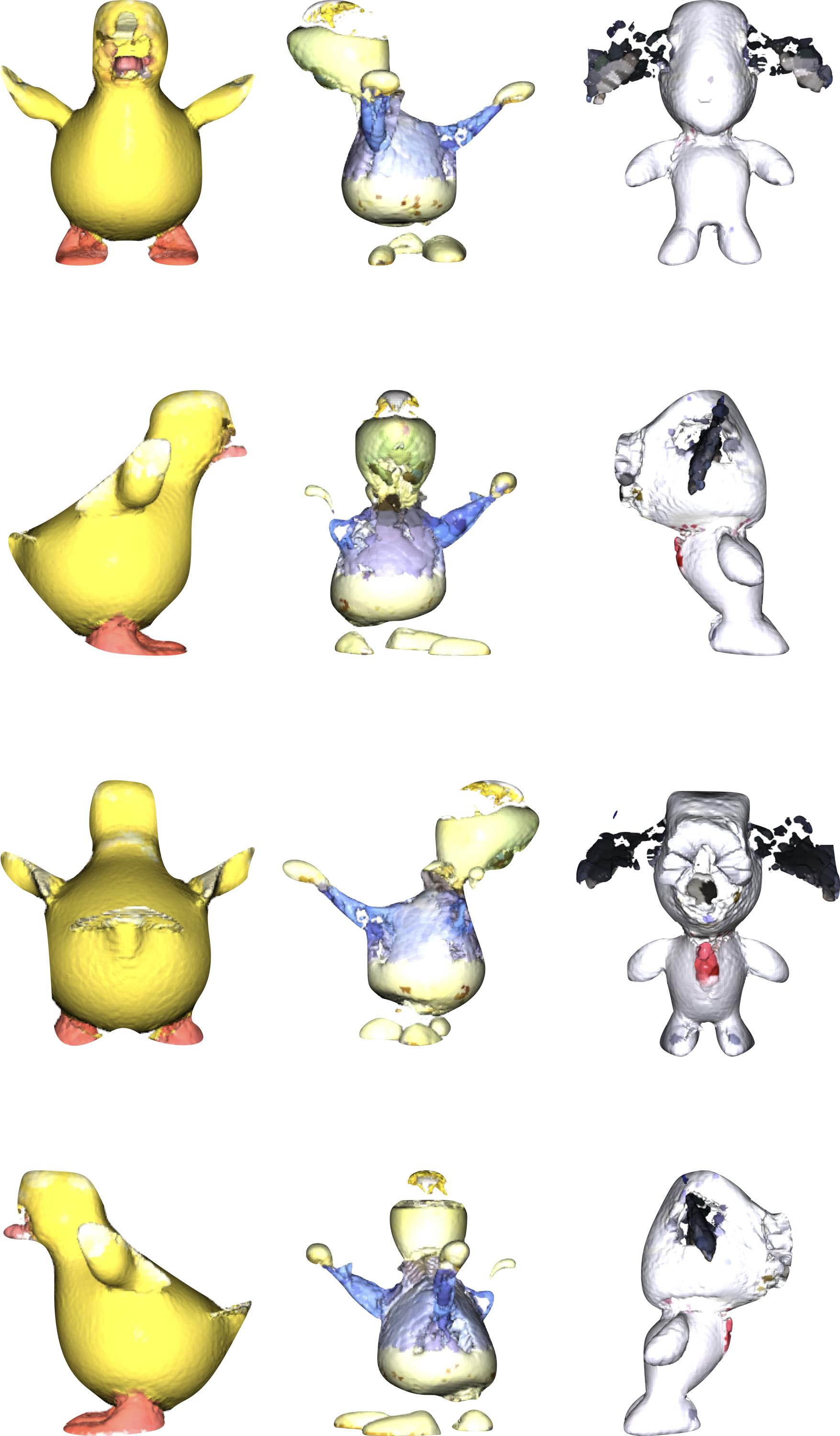} &
    \includegraphics[height=\sz\linewidth]{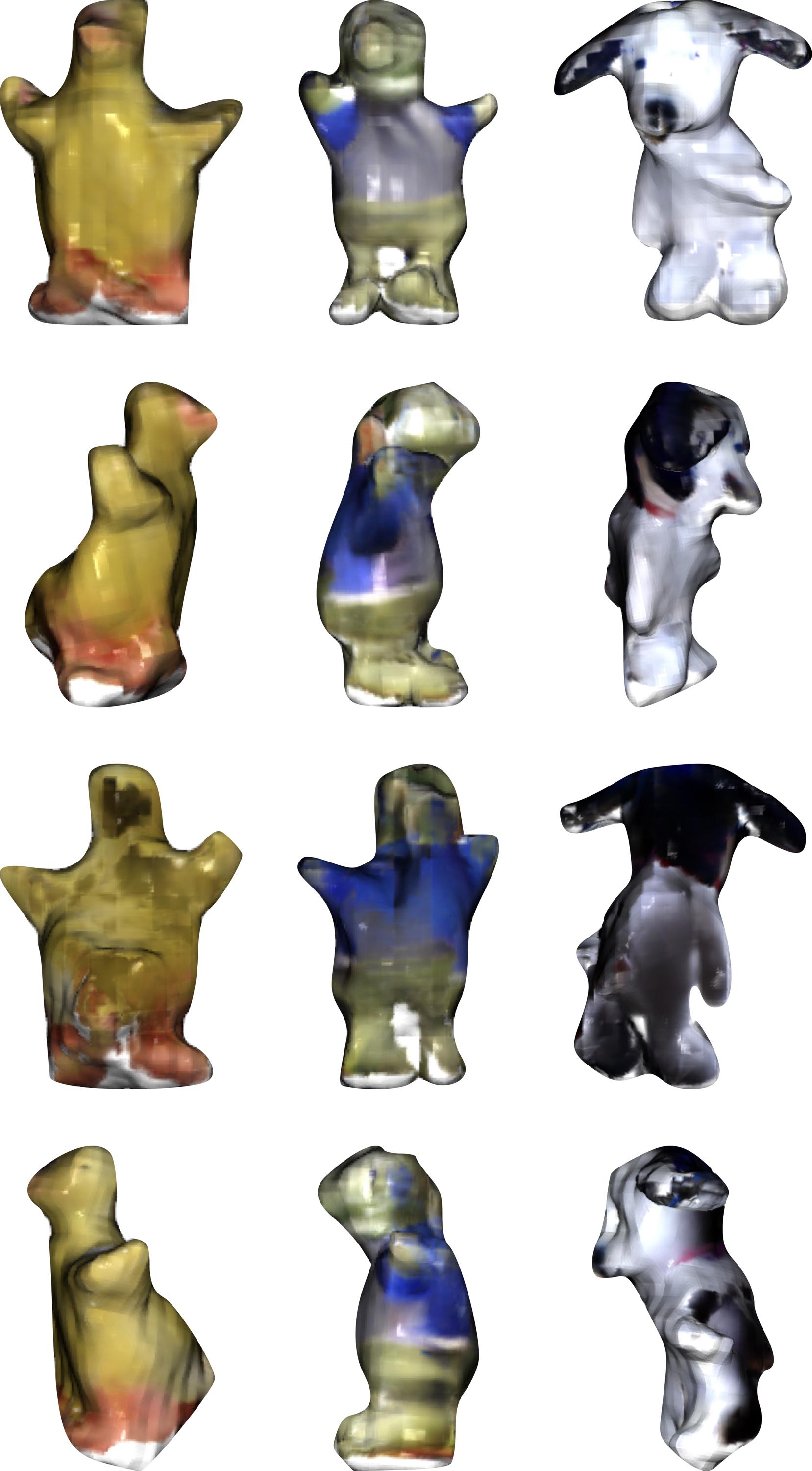} &
    \includegraphics[height=\sz\linewidth]{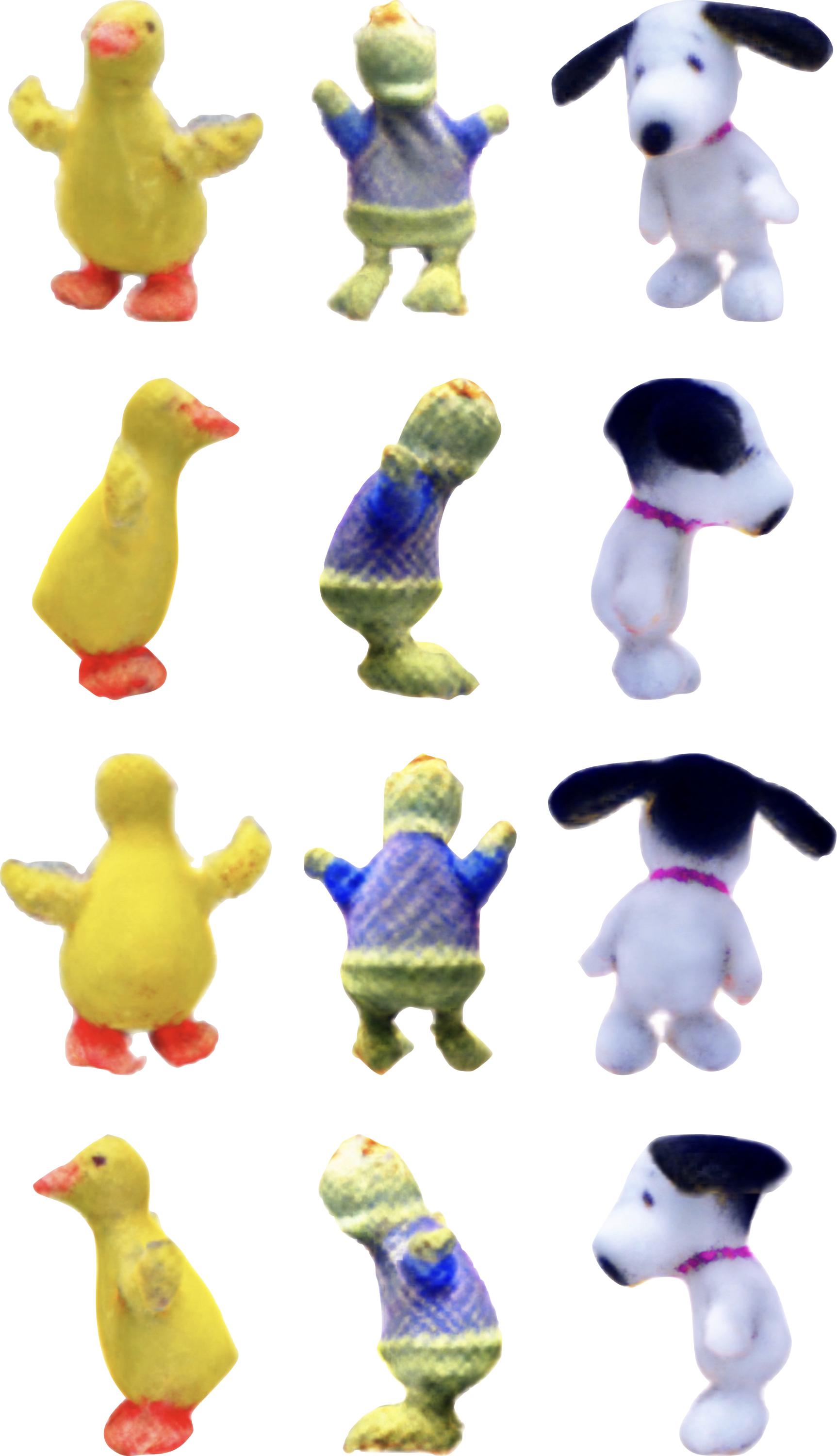} &
    \includegraphics[height=\sz\linewidth]{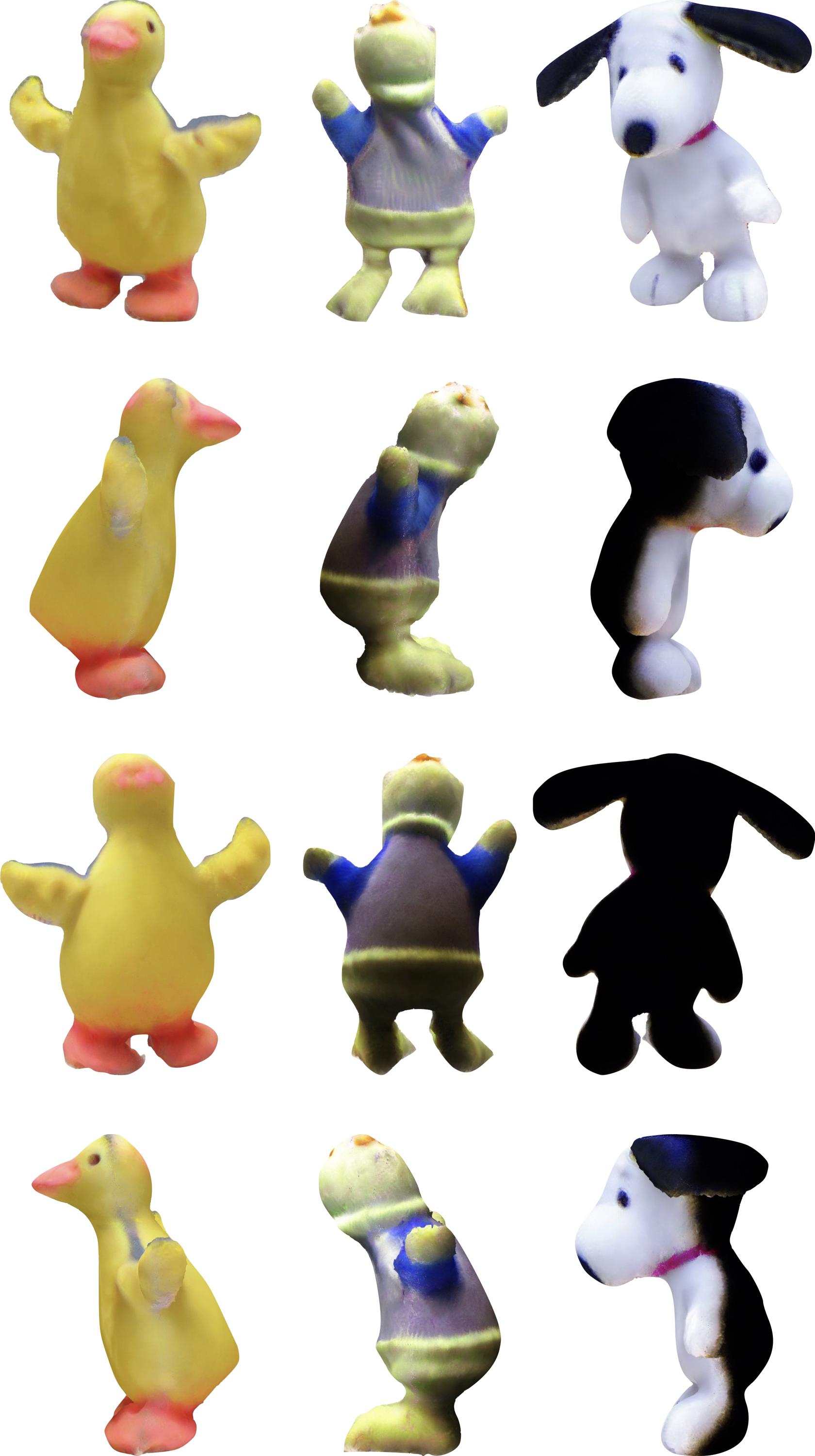}\\

  \end{tabular}
  \caption{\textbf{Comparison of different diffusion priors.} We mainly compare: 1) Point-E~\cite{nichol2022point}: Point cloud-based diffusion model, feed-forward generation 2) One-2-3-45: Generalizable neural surface reconstruction with Zero-1-to-3~\cite{liu2023zero1to3}, feed-forward generation 3) Zero-1-to-3$^\star$~\cite{liu2023zero1to3}: we use Stable-Dreamfusion~\cite{stabledreamfusion2023} repository to perform image-to-3D with SDS from Zero-1-to-3, denoted as Zero-1-to-3$^\star$~\cite{liu2023zero1to3}. The coarse stage uses NeRF with Zero-1-to-3~\cite{liu2023zero1to3} for optimization. The fine stage uses DMTet~\cite{shen2021dmtet} with Zero-1-to-3~\cite{liu2023zero1to3} for optimization.}
  \label{fig:supp_diff_compare}
  \vspace{-4pt}
\end{figure*}

\begin{figure*}[t]
  \centering
  \setlength{\tabcolsep}{10pt}
  \newcommand{\sz}{0.075}
  \begin{tabular}{ccc}
    Reference & Zero-1-to-3$^\star$~\cite{liu2023zero1to3} & Ours\\
    \makecell{\includegraphics[height=\sz\linewidth]{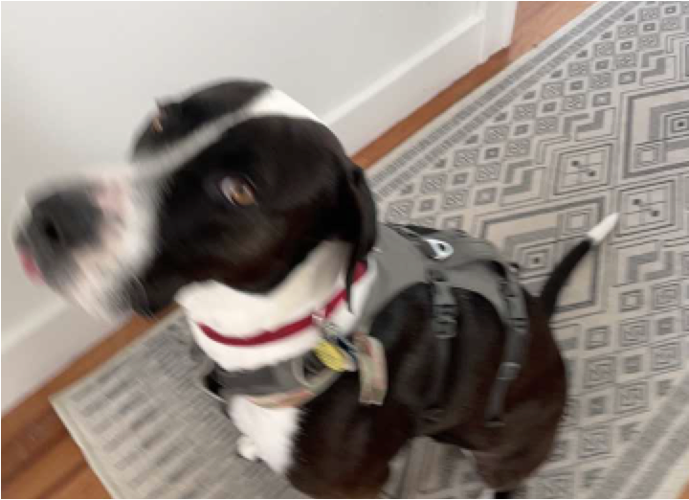}}&
    \makecell{\includegraphics[height=\sz\linewidth]{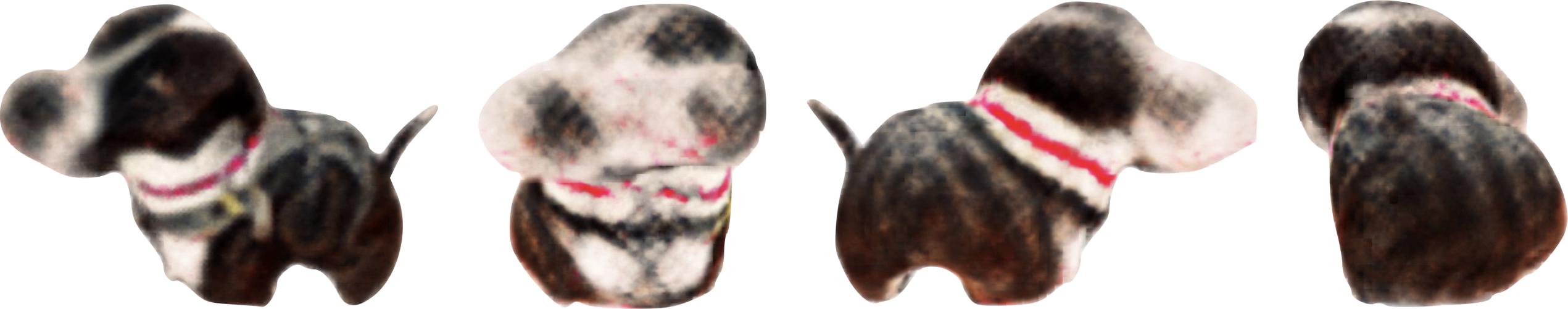}}&
    \makecell{\includegraphics[height=\sz\linewidth]{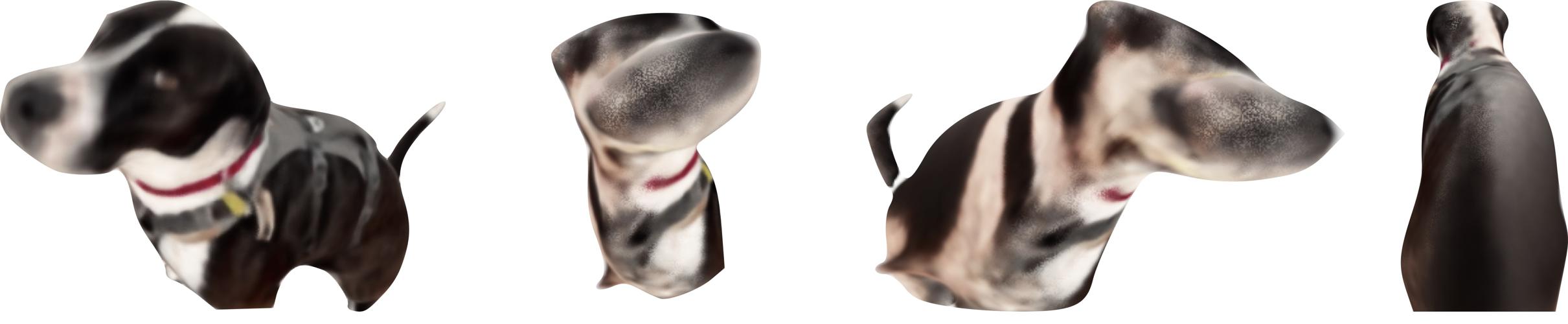}}\\

    \makecell{\includegraphics[height=\sz\linewidth]{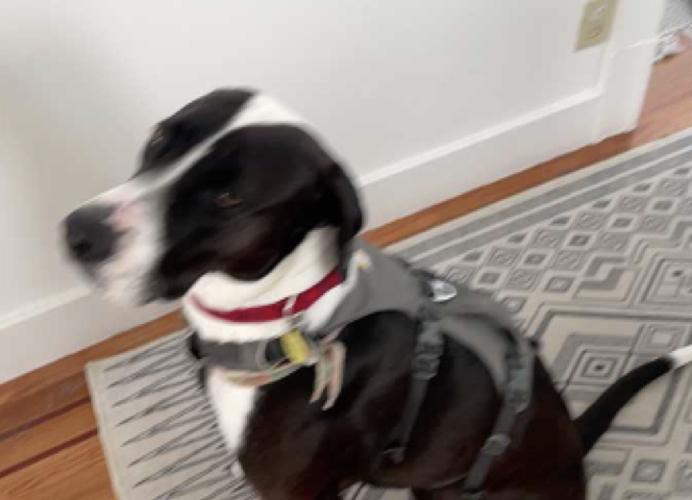}}&
    \makecell{\includegraphics[height=\sz\linewidth]{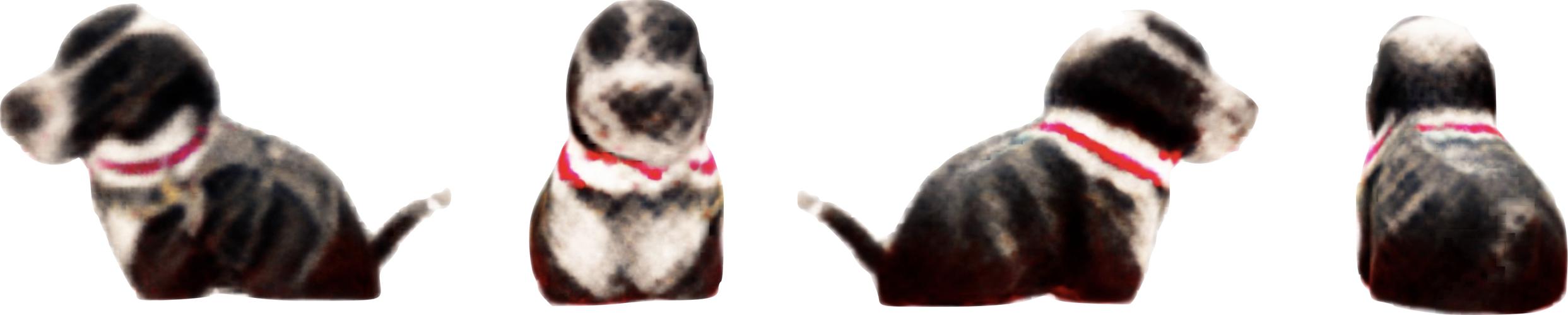}}&
    \makecell{\includegraphics[height=\sz\linewidth]{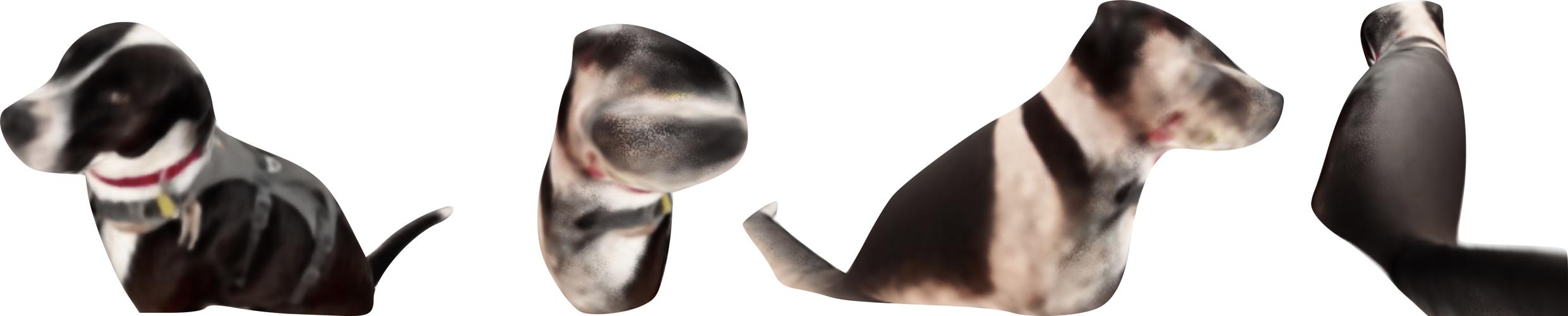}}\\
    
    \makecell{\includegraphics[height=\sz\linewidth]{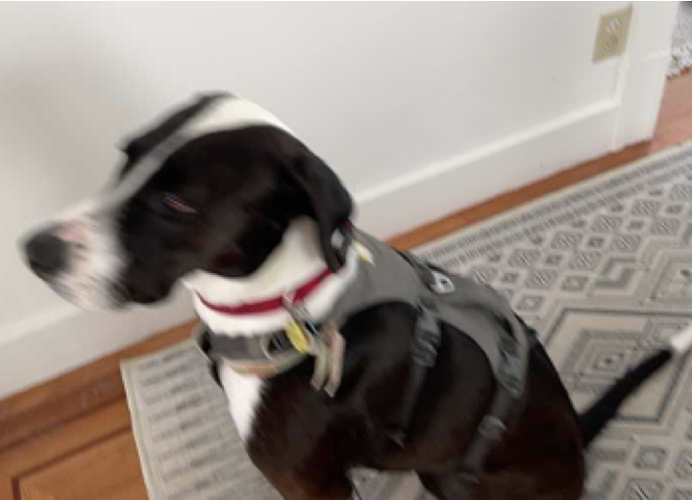}}&
    \makecell{\includegraphics[height=\sz\linewidth]{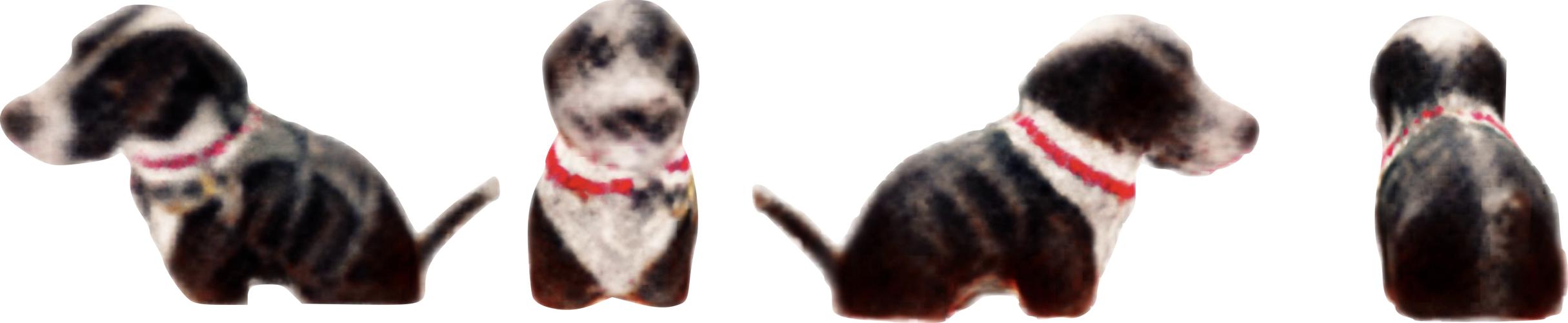}}&
    \makecell{\includegraphics[height=\sz\linewidth]{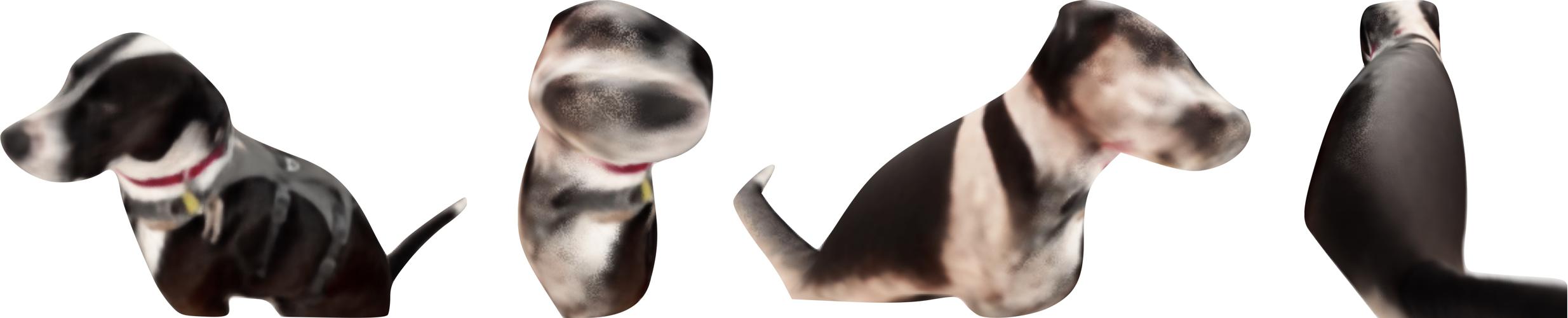}}\\
  \end{tabular}
  \caption{\textbf{Failure case analysis.} We showcase the limitations of \papername{}. The inherent scale ambiguity and challenging real-world scenarios like motion blur and complicated target object pose can hinder the performance of RGB-based diffusion models such as Zero-1-to-3$^\star$\cite{liu2023zero1to3}, causing undesired artifacts like the Janus effect. \papername{} inherits the same limitations but could achieve better results thanks to the leverage of temporal information from the video sequence and regularization on the canonical shape.
  }
  \label{fig:supp_failure}
  \vspace{-4pt}
\end{figure*}
\subsection{Choices of Different Diffusion Priors}
We present a comparative analysis of the generation quality of various diffusion priors in Fig.~\ref{fig:supp_diff_compare}.
Point-E~\cite{nichol2022point} is a diffusion model based on point clouds with feed-forward generation capability. However, its generation may exhibit limitations in accurately fitting the original observations and struggle with uncommon objects (e.g. the Frog Prince toy). 
One-2-3-45~\cite{liu2023one} is another feed-forward generation model, which integrates generalizable neural surface reconstruction with Zero-1-to-3~\cite{liu2023zero1to3}, achieving high-quality image-to-3D generation on synthetic images with remarkable speed. Nevertheless, real-world images often introduce challenges such as high-frequency noise and diverse illumination conditions. One-2-3-45 can fail in these real-world scenarios, and result in artifacts like shadows, inconsistent geometry, missing details, etc.
Zero-1-to-3$^\star$~\cite{stabledreamfusion2023} is another line of work that performs SDS using the Zero-1-to-3~\cite{liu2023zero1to3}. The test-time optimization can effectively get rid of the inconsistent prediction generated by Zero-1-to-3 and result in a coherent geometry (See Fig.~\ref{fig:supp_diff_compare}). Thus, we prefer knowledge distillation from Zero-1-to-3~\cite{liu2023zero1to3} over other feed-forward generation models.

\subsection{Canonical Space Regularization}
We provide more details about the ablation experiments on canonical space regularization. Recall the points used for canonical space regularization in Eq.~\red{12} of our main paper:
\begin{equation}
    \begin{aligned}
    \rvx^\prime_{\mathrm{reg}} = \{\rvx_t, T(\phi(\rvx_t), \gV_t(t)) \}.
    \label{eq_supp:cano_ref}
    \end{aligned}
\end{equation}

\noindent
Note that $\rvx_t$ are sampled directly in the observation space with the deformation network being shortcutted. To encourage the local smoothness of the SDF gradient, a small perturbation $\delta\rvx_t$ is applied to the $\rvx_t$:
\begin{equation}
    \begin{aligned}
    \tilde{\rvx}^\prime_{\mathrm{reg}} = \{\rvx_t+\delta\rvx_t, T(\phi(\rvx_t+\delta\rvx_t), \gV_t(t)) \}.
    \label{eq_supp:cano_ref}
    \end{aligned}
\end{equation}

\noindent
The difference between the gradient of those sets of points $\norm{\nabla_s(\rvx^\prime_{\mathrm{reg}}) - \nabla_s(\tilde{\rvx}^\prime_{\mathrm{reg}})}^2$ is computed as the regularization loss $\gL_{cano}$. 
This loss can effectively constrain the hyper-dimensional canonical field and prevent trivial or ambiguous solutions (e.g. thin geometry with texture carved in it). 

In the ablation experiments, we also experiment with other two variants, both of which involve the use of the deformation network. As opposed to Eq.~\red{12} of our main paper, the points used for regularization:
\begin{equation}
    \begin{aligned}
    \rvx^\prime = \{\rvx_t+D(\phi(\rvx_t), \gV_t(t)), T(\phi(\rvx_t), \gV_t(t)) \},
    \label{eq_supp:cano_deform}
    \end{aligned}
\end{equation}

\noindent
are sampled in the observation space and then deformed to the canonical space. 
For the perturbation vector, we experimented with \textit{obs. perturb.}: applying the perturbation to the points in the observation space $\rvx_t$ before deforming to the canonical space, and \textit{cano. perturb.}: applying the perturbation to the deformed points in the canonical space $\rvx_t+D(\phi(\rvx_t), \gV_t(t))$ and $T(\phi(\rvx_t), \gV_t(t))$.
We find that performing regularization in both ways can lead to over-smooth geometry and trivial solutions (See Fig.~\red{7} in the main paper).

\subsection{Failure Cases Analysis}
We further analyze the failure cases of \papername{} and the challenges for RGB-based diffusion models. 
We show our result on the challenging \textit{haru} sequence from the iPhone dataset. Zero-1-to-3~\cite{liu2023zero1to3} is trained on each individual reference frame for a reference. The results are shown in Fig.~\ref{fig:supp_failure}.

Real-world video captures often have challenging scenarios such as motion blur and complicated articulated poses of the target object (See the images of a dog in Fig.~\ref{fig:supp_failure}). These challenges coupled with the inherent scale ambiguity of RGB observations can hinder accurate shape fitting in the generation process of RGB-based diffusion models, sometimes resulting in undesired artifacts like the Janus effect (See the head of the dog in the first row of Zero-1-to-3$^\star$\cite{liu2023zero1to3} in Fig.~\ref{fig:supp_failure} and the erroneous beak on the back side of the duck in the third row of Zero-1-to-3$^\star$ (Fine) in Fig.~\ref{fig:supp_diff_compare}).

Our \papername{} also relies on an RGB-based diffusion model and thus also inherits the same challenges and difficulties. However, the leverage of temporal information from the entire video sequence and the implicit regularization of the canonical field can allow \papername{} to alleviate the above-mentioned problems in Zero-1-to-3. For instance, the Janus effect can be eliminated. See the head of the dog in our reconstruction result.

{
    \small
    \bibliographystyle{ieeenat_fullname}
    \bibliography{main}
}
